%% file: ijcai23.tex
\documentclass{article}
\usepackage{ijcai23}

\usepackage{times}
\usepackage{soul}
\usepackage{url}
\usepackage[hidelinks]{hyperref}
\usepackage[utf8]{inputenc}
\usepackage[small]{caption}
\usepackage{graphicx}
\usepackage{amsmath}
\usepackage{amsthm}
\usepackage{booktabs}
\usepackage{adjustbox}

\input{macros}\usepackage{xr}
\title{\modehb: Evolutionary-based Hyperband for Multi-Objective Optimization}

\author{
Noor Awad$^1$\thanks{Equal Contribution}\and
Ayushi Sharma$^1$\footnotemark[1]\and
Philipp Muller$^2$\and
Janek Thomas$^2$\and
Frank Hutter$^1$
\affiliations
$^1$Department  of  Computer  Science,  University  of  Freiburg, Germany\\
$^2$Department of Statistics, University of Munich, Germany\\ 
\emails
\{awad,sharmaa,fh\}@informatik.uni-freiburg.de
}

\begin{document}

\maketitle

\begin{abstract}
    Hyperparameter optimization (HPO) is a powerful technique for automating the tuning of machine learning (ML) models. However, in many real-world applications, accuracy is only one of multiple performance criteria that must be considered. Optimizing these objectives simultaneously on a complex and diverse search space remains a challenging task. In this paper, we propose \modehb{}, an effective and flexible \mo{} (MO) optimizer that extends the recent evolutionary Hyperband method \dehb{}. We validate the performance of \modehb{} using a comprehensive suite of 15 benchmarks consisting of diverse and challenging MO problems, including HPO, neural architecture search (NAS), and joint NAS \& HPO, with objectives including accuracy, latency and algorithmic fairness. A comparative study against state-of-the-art MO optimizers demonstrates that \modehb{} clearly achieves the best performance across our 15 benchmarks.
\end{abstract}

\section{Introduction}
Designing the architecture and tuning hyperparameters for complex machine learning (ML) methods can be costly, time-consuming, and error-prone. 
In recent years, there has been a significant amount of research on methods such as \hpo{} (HPO)~\cite{feurer-automlbook19a} and \nas{} (NAS)~\cite{elsken-jmlr19a} to address these challenges. 
Various methods have been developed to efficiently solve these tasks, however, many of them only focus on optimizing a single objective, such as predictive accuracy. 
In real-world applications, practitioners often need to optimize multiple objectives that may conflict with each other, such as finding a smaller model with minimal drop in predictive performance. 
This can be a difficult task, particularly when deploying models on resource-constrained devices. 
Another omnipresent example is ensuring predictive performance while guaranteeing a certain level of fairness. 
This poses a challenge and significant trade-offs between performance and different fairness notions need to be explored ~\cite{corbett-kdd17}.
Despite efforts to adapt single-objective (SO) optimizers for multi-objective (MO) settings, there have been relatively few studies that have examined the use of these methods for HPO and NAS in complex search spaces.

Because HPO and NAS require a lot of computational resources when executed naively, multi-fidelity (MF) techniques have gained popularity.
These methods exploit cheap approximations, such as evaluations at fewer epochs or on smaller data subsamples, of the objective function to speed up the optimization. 
Successive Halving (\sh{})~\cite{jamieson-aistats16a} is a first simple, yet effective MF method that allocates resources to configurations that perform sufficiently well,
 promoting them to run for larger computational budgets, rather than evaluating poor ones for the maximum resource budget. 
The Hyperband (\hb{})~\cite{li-iclr17a} algorithm instantiates \sh{} with different 
levels of its lowest fidelity,
to guard from sub-optimal choices of fidelity levels. 
While SH and HB are effective in choosing the fidelity level of the configurations they evaluate, they rely on simple random search for choosing those configurations and therefore do not learn over the course of the algorithm.
To address this issue, BOHB~\cite{falkner-icml18a} introduces the use of Bayesian Optimization (BO) via Tree Parzen Estimates (\tpe{})~\cite{bergstra-nips11a}, and \dehb{} \cite{awad-ijcai21a} uses an evolutionary algorithm (EA) based on Differential Evolution (\de{})~\cite{storn-jgo97a} to further improve the search at each budget level.

A SO optimizer can be extended to handle MO by using (1) scalarization technique (i.e., optimizing a linear combination of multiple competing objectives to a single objective point) or (2) non-dominated sorting (NDS) (i.e., approaches that take the geometry of the whole Pareto front into account). 
Although scalarization is a simple convenient technique in MO, it has multiple shortcomings: (1) it captures only restricted pareto shapes, (2) linear scalarization can cause failures for non-convex pareto fronts, (3) it encourages a single direction exploration, (4) it fails to recover the full pareto front if the weights are not chosen carefully and (5) it is greatly affected by re-scaling the values of competing objectives.

To mitigate the aforementioned issues of scalarization methods, 
in this work we extend \dehb{} to optimize multiple objectives by the use of NDS. 
To summarize, our main contributions are:
\begin{enumerate}
    \item We extend the \hb{} component in \dehb{} to MO considering two variants: (1) NSGA-II and (2) EpsNet.
    \item We adapt the evolutionary search component in \dehb{} to handle MO by incorporating NDS and dominated hypervolume contribution in DE selection strategy to  approximate the desired trade-offs between the objectives. 
    \item We comprehensively evaluate our two \modehb{} variants on a wide range of 15 MO problems, and demonstrate that they establish the new state of the art in MO compared to many state-of-the-art baselines.  
\end{enumerate}

The paper proceeds as follows: Section~\ref{sec-related-work} presents related work. 
Section~\ref{sec-background} provides background on MO, a summary of MO evolutionary algorithms (EAs), and an overview of \dehb{}. 
Section~\ref{sec-modehb} presents our proposed method, \modehb{}.
Section~\ref{sec-experiments} empirically evaluates \modehb{} on a broad range of 15 benchmarks from three benchmark families, showing that it defines a new state-of-the-art performance. 
Finally, Section~\ref{sec-conclusion} concludes the paper.

\section{Related Work}
\label{sec-related-work}
BO and EAs are popular choices for solving MO HPO problems. MO EAs are usually based on Pareto dominance (e.g., NSGA-II~\cite{deb-tevc2002}), decomposition (e.g., MOEA/D~\cite{zhang-tevc2007}) or indicators (e.g., SMS-EMOA~\cite{emmerich-emc2005})~\cite{emmerich-nc18}.
MO BO algorithms are usually based on scalarization (e.g., ParEGO~\cite{knowls-evoco06a}), aggregating acquisition functions (e.g., SMS-EGO~\cite{ponweiser-ppsn08}), multiple acquisition functions (e.g., MultiEGO~\cite{jeong-cec05}) or information theoretic (e.g., PESMO~\cite{hernandez-icml16a}).
A recent review of these methods and their use in MO HPO can be found in \cite{karl-arxiv22}.
However, these methods do not generally support multi-fidelity evaluations. 
Recent work introduces the first MO \hb{} method by using a simple scalarization method~\cite{Schmucker-metalearn20} or NDS~\cite{schmucker2021multi} to overcome the 
shortcomings of scalarization methods.
The NDS approach has been used to jointly optimize hyperparameters and hardware configuration~\cite{salinas-arxiv-21}.

\section{Background}
\label{sec-background}
\subsection{Multi-objective Optimization}
\label{subsec-MO}
Let $f : \Lambda \to \R^n$ be a function over the search space $\Lambda$.
A multi-objective optimization problem aims to minimize a function $f : \Lambda \to \R^n$ over the search space $\Lambda$ where $n$ is the number of the objectives. 
\begin{align}\label{eq:multi_obj}
  \min_{\lambda \in \Lambda} f(\lambda) = \min_{\lambda \in \Lambda} \left( f_1(\lambda), f_2(\lambda), \ldots, f_n(\lambda) \right) \ .
\end{align}
In our application this would represent $n$ different performance measures of a machine learning model, configured by the hyperparameter vector $\lambda$.
One objective is the estimated generalization error and others can, among other things, be related to model size, inference time and fairness.
Given two configurations $\lambda^{(1)}, \lambda^{(2)} \in \Lambda$, we say $\lambda^{(1)} \succ \lambda^{(2)} $ if $\lambda^{(2)}$ is dominated by $\lambda^{(1)}$, that is, $f_{i}(\lambda^{(2)}) \geq f_{i}(\lambda^{(1)}) \forall i \in [n] $ and $f_{j}(\lambda^{(2)}) > f_{j}(\lambda^{(1)})$ for at least one $j \in [n] $.
Configurations $\lambda^*$ that are not dominated by any other candidates are called Pareto optimal, and the set of Pareto optimal configurations forms the Pareto front.
Any MO algorithm aims to find the true Pareto set or an approximation set of non-dominated solutions. 
In practice, it is rarely possible to compute the true Pareto set due to the complexity of the search space $\Lambda$, potential noise in the objective functions and the difficulty in optimizing multiple competing objectives.

Hypervolume (HV) is one of the most commonly used quality indicators in MO~\cite{Zitzler-ppsn98,knowles2003bounded}. 
The hypervolume indicator $\mathcal{H}(A)$ of A is the measure of the region weakly dominated by A and bounded above by a reference point $r$:
\begin{align}\label{eq:hv}
    \mathcal{H}(A) = \Lambda({\lambda \in \mathcal{R} ^d | \exists p \in A: p\succeq  \lambda \wedge \lambda \succeq r  }),
\end{align}
where $\Lambda$ is the Lebesgue measure. HV has been shown to promote the convergence towards the Pareto front, as well as to ensure a representative distribution of points along the front. 
Maximzing HV is equivalent to identifying the Pareto set, and with a limited number of configurations, maximizing HV metric results in subsets of the Pareto front that are evenly distributed~\cite{beume2007sms,guerreiro2021hypervolume}. 

\subsection{Multi-objective Evolutionary Algorithms}
\label{subsec-MOEAs}
Non-dominated sorting (NDS) is a widely used method in \mo{} EAs~\cite{srinivas1994muiltiobjective}. 
It is used to sort candidates based on their non-domination level, which is determined by partitioning the objective space into fronts. 
NDS partitions an objective space set into fronts $F_{1} \prec F_{2} \prec \cdots \prec F_{n}$ such that a configuration $\lambda^1 \in F_{i}$ outperforms another candidate $\lambda^2 \in F_{j}$ with respect to all objectives if $i < j$. 
We can easily compute the partitioning by computing all non-dominated points $F_{1}$, removing them, then computing the next layer of non-dominated points $F_{2}$, and so on. 
 
The candidates within the same front are further sorted by their crowding distance~\cite{deb2002fast}, which is calculated by comparing the distance between a candidate and its two nearest neighbors for each objective function. To calculate Crowding distance, we follow the following steps for every objective $f_{j}$: \textbf{(i)}~Sort points by $f_{j}$. \textbf{(ii)}~Normalize scores to [0,1]. \textbf{(iii)}~The boundary solutions (solutions with smallest and largest function values) are assigned an infinite distance value. Therefore, they are always selected. \textbf{(iv)}~For each point a distance score, which is the distance between its 2 next-neighbors w.r.t. the sorting of $f_{j}$ is assigned. The overall crowding-distance value is calculated as the sum of individual distance values corresponding to each objective. 
 Another approach to sort candidates within the same front is the EpsNet exploration strategy~\cite{salinas2021multi}. This strategy starts by removing the first point from the "front" set (which represents all points on the front that we aim to  sort) and placing it in an empty set, referred to as the "ranked" set. Subsequently, the remaining points in the "front" set are iteratively selected based on their Euclidean distance from the points already in the "ranked" set. Specifically, the strategy selects the point that has the highest Euclidean distance with respect to its closest point in the "ranked" set. The selected point is then removed from the "front" set and placed into the "ranked" set. This process continues iteratively until all points have been ranked. By selecting the point that has the highest distance to its closest point in the "ranked" set, the strategy ensures that the selected candidates are as dissimilar as possible. This can be useful for exploring different regions of the solution space and for identifying trade-offs between different objectives.

\subsection{Overview of DEHB}
\label{subsec-overview-dehb}
\begin{figure}
\centering
\includegraphics[width=0.9\columnwidth]{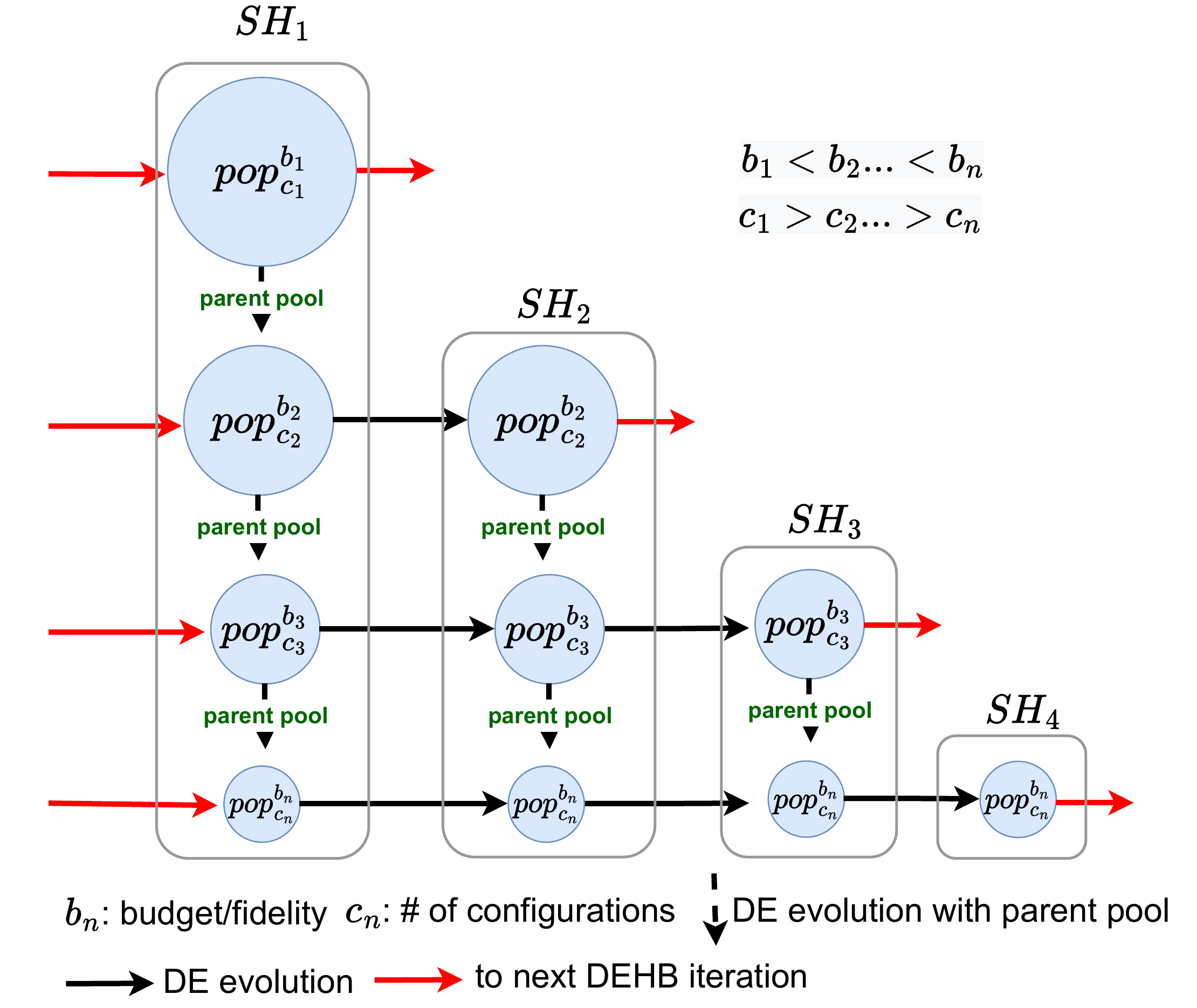}
\caption{DEHB iteration}
\label{fig:DEHB-iteration}
\end{figure}

\dehb{} is a recently proposed multi-fidelity evolutionary optimizer that uses a new amalgam of Differential Evolution and Hyperband for HPO~\cite{awad-ijcai21a}. 
\dehb{} fulfills the desiderata of a good HPO optimizer such as: (1) conceptually simple (2) computationally cheap (3) strong final and anytime performance and (4) effectively takes advantage of parallel resources. 
Adding to that, \dehb{} is very effective in optimizing high dimensional and discrete search spaces.
A key design principle of \dehb{} is to share information across the runs it executes at various fidelities. 
Unlike \hb{} that only prompt the top performing configurations from lower fidelity to higher one, \dehb{} maintains a sub-population for each fidelity and uses an evolutionary search based on \de{}. 
Each sub-population is evolved and updated using mutation, crossover and selection operations. 
Figure \ref{fig:DEHB-iteration} illustrates the framework of a \dehb{} iteration where $\eta=3$ \textit{minimum fidelity $b_{min}=3$} and \textit{maximum fidelity $b_{max}=27$}.
The first \sh{} bracket of \dehb{} iteration initializes all the sub-populations randomly and follows the \sh{} subroutine as in vanilla \hb{}.
Starting from the second iteration, \dehb{} performs a modified \sh{} using DE evolution.
For each \sh{} iteration (top to bottom), \dehb{} collects the top performing configurations for each fidelity level in a parent pool
which is used to select the parents for the mutation instead of using the current sub-population itself. 
By doing so, the DE evolution can still incorporate information from the good performing region w.r.t. the lower fidelity. When the size of a sub-population is small \footnote{if the population size is less than 3 (i.e., minimum number of parents) needed to perform the used \textit{rand1} mutation strategy in DE} which usually occurs on full fidelity level, a global population that consists of all the configurations from all the sub-population is used to select additional parents to perform mutation.
   
The crossover then combines each individual and its corresponding mutant vector to generate the final offspring. 
Next, the selection operation updates the sub-population on the current fidelity if the new offspring is better that its parent. 
The sub-populations (right to left) are forwarded and evolved from a \sh{} iteration to a next one. 
The final evolved sub-populations in each \sh{} are then forwarded to start a new \dehb{} iteration.

\section{\modehb{}: Algorithm Description}
\label{sec-modehb}
Multi-objective EAs usually navigate the objective space and optimize the Pareto front using NDS as discussed in Section~\ref{subsec-MOEAs}. 
To leverage suitability and flexibility of \dehb{} for the MO domain, we follow a similar approach and apply two changes. 
First, we follow a similar approach as proposed in~\cite{Schmucker-arxiv21} to adapt the \hb{} component for selecting the top-performing configurations in MO domain using NDS with the use of NSGA-II or Epsnet to rank the Pareto fronts resulting in two variants. 
Second, we extended the evolutionary search using DE to MO based on NDS and dominated HV. 
The following subsections explain these changes in more detail. 

\subsection{Multi-objective Hyperband}
\label{sub-sec-MOHB}
The single-objective \hb{} starts by instantiating a \sh{} subroutine by making use of the input parameters (i.e., minimum fidelity, maximum fidelity and $\eta$) that used to trade-off the number of configurations and fidelity. The first iteration of \sh{} samples $N$ random configurations with a small initial budget, evaluates them and forwards the top $1/\eta$ configurations to the next higher budget. This process is repeated until the highest budget is reached. Subsequent \sh{} iterations promote the top $1/\eta$ configurations from the lower fidelity to the parent pool which is further used in DE evolution. To select the top-performing configurations, we adapt \hb{} for the MO domain by employing a methodology similar to that presented in~\cite{schmucker2021multi}. This approach is based on NDS to rank the configurations in the lowest budget by taking into account the global geometry of the Pareto front approximations in two stages. 
Referring to Figure \ref{fig:DEHB-iteration}, for each \sh{} bracket we first rank the population into multiple fronts using NDS. 
Then, to rank the candidates in the same front (i.e., to choose a subset of the population from a specific front) we employ one of two strategies: 
The first strategy is based on NSGA-II (we dub it \modehbnsgaii{}) which uses the crowding-distance to select the candidates with the lower density in the objective space. 
The second strategy, EpsNet (giving rise to \modehbepsnet{}), aims to iteratively select candidates by maximizing the distance between them and the previously chosen configurations in the objective space. 
Both of these strategies aim to select a sparse set of candidates that effectively represents the front, with the goal of promoting diversity of the parent pool when we later use them to perform the mutation in DE search to find good solutions.

For example, in the $SH_1$ bracket in Figure~\ref{fig:DEHB-iteration}, we rank the population, $pop_{c_1}^{b_1}$, into multiple fronts using NDS. 
Let us assume $pop_{c_1}^{b_1}$ is divided into three fronts $F_1$, $F_2$ and $F_3$.
Next, assuming the number of candidates to be promoted to higher fidelity (which is determined by \hb{}) is greater than the number of candidates in $F_1$ but smaller than $F_2$, we select all the candidates from $F_1$ and select the remaining ones from $F_2$ by either crowding-distance or EpsNet.

\subsection{Multi-objective DE}
\label{subsec-MODE}
In \dehb{}, the configurations at each fidelity level form a sub-population and are evolved using an evolutionary search based on the following \de{} operations: mutation, crossover and selection. 
For each \sh{} iteration (top to bottom in Figure \ref{fig:DEHB-iteration}), the parent pool is used to select three random parents to perform the mutation operation (i.e., \textit{rand1} mutation strategy~\cite{awad-ijcai21a}) and generate a new mutant configuration. 
We then apply the binomial crossover between the generated mutant and the target configuration to generate the final offspring.
Next, the offspring is evaluated and compared to its target configuration to select which one performs better to replace the other.
To adapt DE to MO, we extend the selection operation to handle multiple objectives as we present in Algorithm~\ref{alg:MO-DE}. 
We utilize the global population in \dehb{} that contains the most recent sub-populations that have evolved across all fidelities. We incorporate new offspring into it and sort it according to NDS, resulting in multiple fronts. The front rankings of both the target and offspring are then determined, which helps in positioning them in the objective space with their respective trade-offs.  
If the offspring is on a lower ranked front than the target, it replaces the target in the sub-population. 
If the target is on a lower ranked front, the sub-population is not updated. 
If they are on the same front, then the one with the least dominated hypervolume in the last front is chosen to be replaced by the offspring, as it has the least contribution in approximating the desired trade-offs.

\begin{algorithm}[!ht]
\KwIn{{}\\
  {$\rho$ $\leftarrow$ parent\_pool; - Parent pool from lower fidelity\\}
  {$pop$ $\leftarrow$ population; - Population for current fidelity \\}
  {$global\_pop$ $\leftarrow$ - Population across all fidelity \\}
  }
  \KwOut{{}Return evolved popluation }
	\For{$parent$ in $pop$}{
		$mutant$ = generate\_mutant($\rho$);\\
		$offspring$ $\leftarrow$ crossover($mutant$, $parent$);\\
		$f$ = evaluate($offspring$); \\
		$C$ $\leftarrow$ MO-selection($pop$,$parent$,$offspring$); \\
        update population($C$, $offspring$, $global_pop$)
	}
    
    \SetKwFunction{FMain}{MO-selection}
    \SetKwProg{Fn}{Function}{:}{}
    \Fn{\FMain{$global\_pop, parent, offspring$}}{
        $[F_1,...,F_m] \xleftarrow{} NDS(global\_pop)$\\
        $R\_parent, R\_offspring$ = get\_front\_ranking($parent, offspring, [F_1,...,F_m]$)\\
        \uIf {$R\_parent > R\_offspring$} {
            return $parent$
            }
        \uElseIf {$R\_parent < R\_offspring$} {
            return $offspring$
            }
        \uElse{
            $least\_hv\_contributor$ = minimum\_HV($F\_m$)\\
            return $least\_hv\_contributor$
            }
    }
    
	\Return $pop$
	\caption{MO-DE}\label{alg:MO-DE}
\end{algorithm}


\section{Experiments}
\label{sec-experiments}
We evaluated \modehb{} on a broad collection of MO problems with a total of $15$ benchmarks from diverse domains: (i) NAS (9 benchmarks), (ii) joint NAS \& HPO (5 benchmarks) and (iii) algorithmic fairness (1 benchmark). 
In the following subsection we explain our benchmarks, experimental setup and results in detail.

\subsection{Benchmarks}
\label{sec-benchs}
\subsubsection{Neural Architecture Search}
\label{sub-sec-NAS-bench}
The complexity of ML models has led to a widespread use of NAS to find efficient architectures~\cite{elsken-jmlr19a}. 
However, NAS methods often focus solely on achieving the optimal performance and do not take into account other important objectives such as latency. 
To address this, we conduct experiments on NAS-Bench-101~\cite{ying2019bench}, NAS-Bench-1shot1~\cite{zela2020bench} and NAS-Bench-201~\cite{dong2020bench}, which are 9 tabular benchmarks. 
Our experiments aim to optimize for both validation accuracy and model size. 
For more information about these benchmarks, refer to \hpobench{} suite~\cite{eggensperger-neuripsdbt21a}.

\subsubsection{Joint NAS \& HPO}
\label{subsec-bench-NASHPO}
While there has been extensive research on optimizing architecture and model hyperparameters separately, few studies have investigated optimizing them together. 
In this experiment, we use two raw benchmarks. The first one, proposed by~\cite{izquierdo2021bag}, involves tuning Convolutional Neural Networks (CNNs) composed of a variable number of fully-connected layers and a 15D mixed and hierarchical search space. 
These networks are trained on the Oxford-Flowers dataset~\cite{Nilsback08} and Fashion-MNIST~\cite{xiao2017fashion}. For these two datasets, we optimize validation accuracy and model size.
For more information on the datasets and search spaces, refer to~\cite{izquierdo2021bag}. The second benchmark uses three surrogate benchmarks~\cite{zela-iclr22a} from the recently introduced JAHS-Bench-201 suite~\cite{bansal2022jahs} (CIFAR-10, Colorectal-Histology and Fashion-MNIST), and we optimize validation accuracy and latency.
 
\subsubsection{Algorithmic Fairness}
\label{subsec-bench-fairness}
We adopt the experimental setup from~\cite{Schmucker-arxiv21} and train a multi-layer perceptron (MLP) to optimize both predictive accuracy and fairness. 
Specifically, we use the fairness metric of difference in statistical parity (DSP) which measures the absolute difference in predicted outcome between two subgroups distinguished by a protected attribute (i.e., sex). 
The MLP is trained on the Adult dataset~\cite{kohavi1996scaling}. 
For more information about this raw benchmark, refer to~\cite{Schmucker-arxiv21}. 

We provide more details about the benchmarks in Appendix B.

\subsection{Experimental Setup}
\label{sub-sec-exp-setup}
\textbf{Evaluation protocol} 
We ran our experiments for $10$ repetitions for each benchmark using different seeds. 
We set the wallclock runtime limit to $24$h for raw benchmarks (Flower, Fashion and Adult) and $4$h for surrogate and tabular benchmarks.
For all benchmarks, we further introduce a limit on the number of target algorithm executions,
determined by $\big\lceil 20 + 80 \times \sqrt{\text{\#HPs}}\big\rceil$, similar to~\cite{pfisterer-automl22} where HPs denotes the number of hyperparameters. 
We report the difference in log HV from the best possible Pareto set to the one currently observed by an optimizer: $\text{LogHVDiff}(\text{HV})~=~log_{10}(\text{HV} - \text{HV}_{emp})$ where $\text{HV}_{emp}$ is the normalized HV of all observed configurations over all the runs. 

\textbf{Hardware} 
For all surrogate and tabular benchmarks, we run  experiments on a compute cluster equipped with Intel(R) Xeon(R) Gold 6242 CPUs @ 2.80GHz. 
For the raw benchmarks, we ran all jobs on a compute cluster equipped with NVIDIA GeForce RTX 2080Ti.

\textbf{Baselines} In our comparison, we compare our two proposed variants (\modehbnsgaii{} and \modehbepsnet{}) against state-of-the-art baselines. 
We use two \mo{} BO optimizers: \qnparego{}~\cite{daulton-neurips20a} and SMAC~\cite{lindauer2022smac3} and three \mo{} EAs: \nsgaiii{}~\cite{blank2019investigating}, \agemoead{}~\cite{panichella2019adaptive} and \moead{}~\cite{zhang2007moea}. 
We further add random search (\RS{}) as a baseline. 
We use the default settings to run these baselines and report that as well as more information about them in Appendix A. 

\subsection{Results}
\label{subsec-resuls}

\begin{figure*}[ht]
  \begin{center}
  \begin{tabular}{l@{}l@{}l}
        \includegraphics[trim=0 1cm 1cm 0cm   ,clip,width=0.315\textwidth]{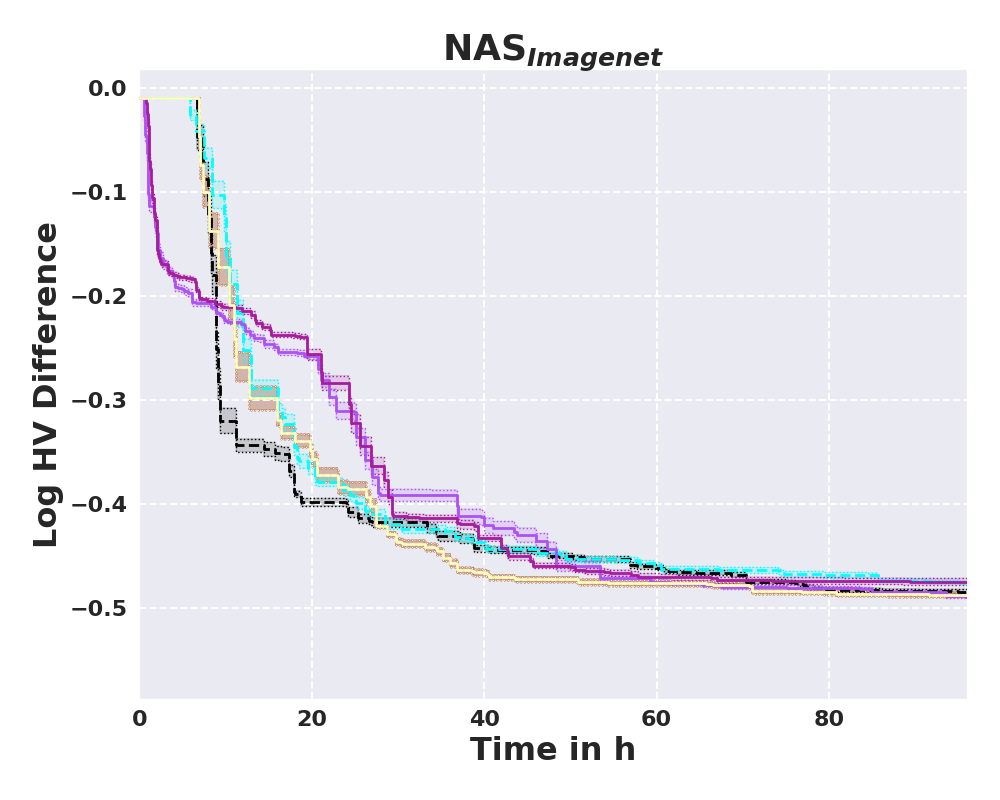}
        & \includegraphics[trim=0.8cm 1cm 0cm   0cm,clip,width=0.315\textwidth]{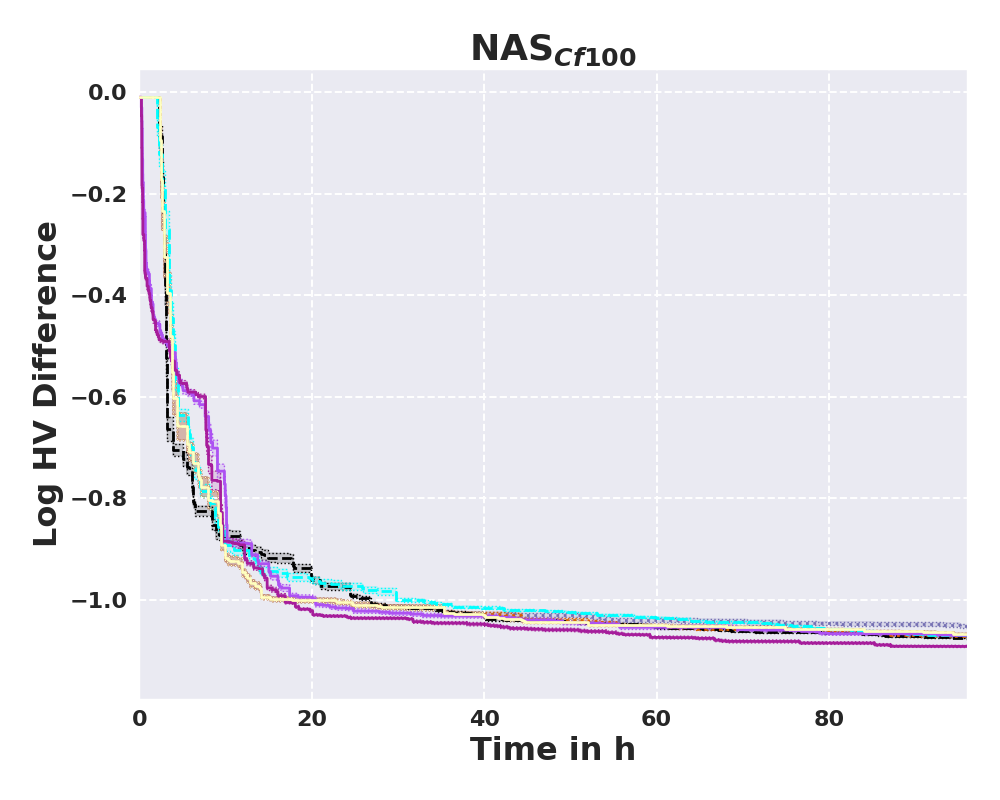}
       & \includegraphics[trim=0.8cm 1cm 0cm   0cm,clip,width=0.315\textwidth]{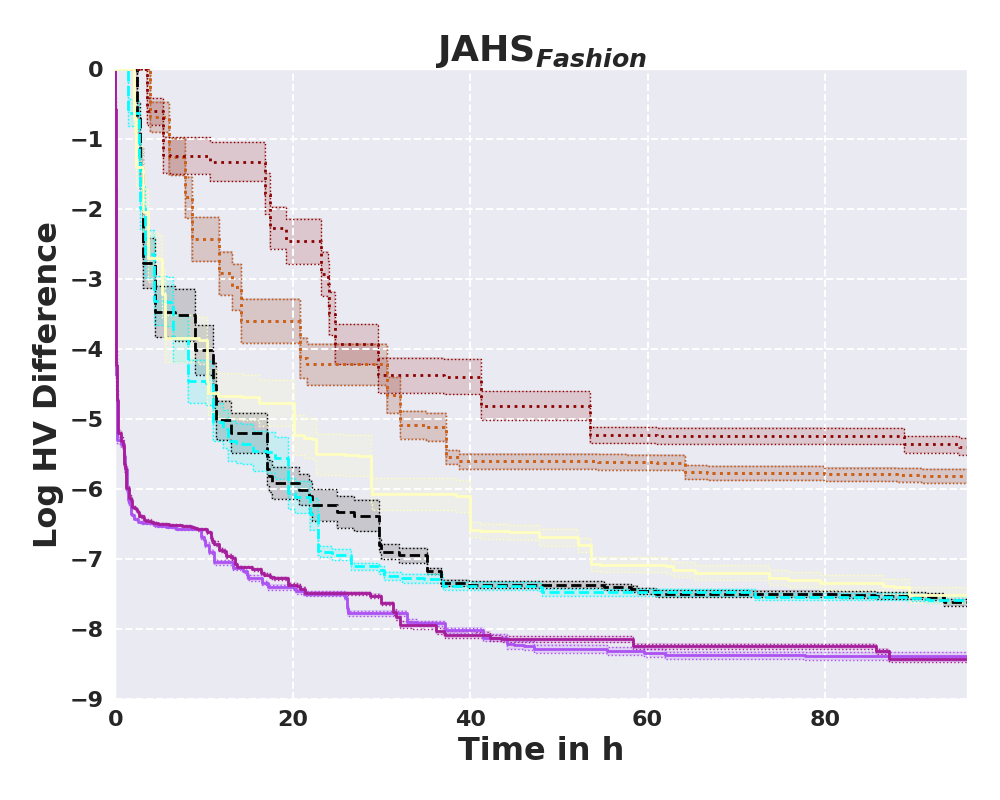}\\
       \includegraphics[trim=0 1cm 0cm   0cm,clip,width=0.33\textwidth]{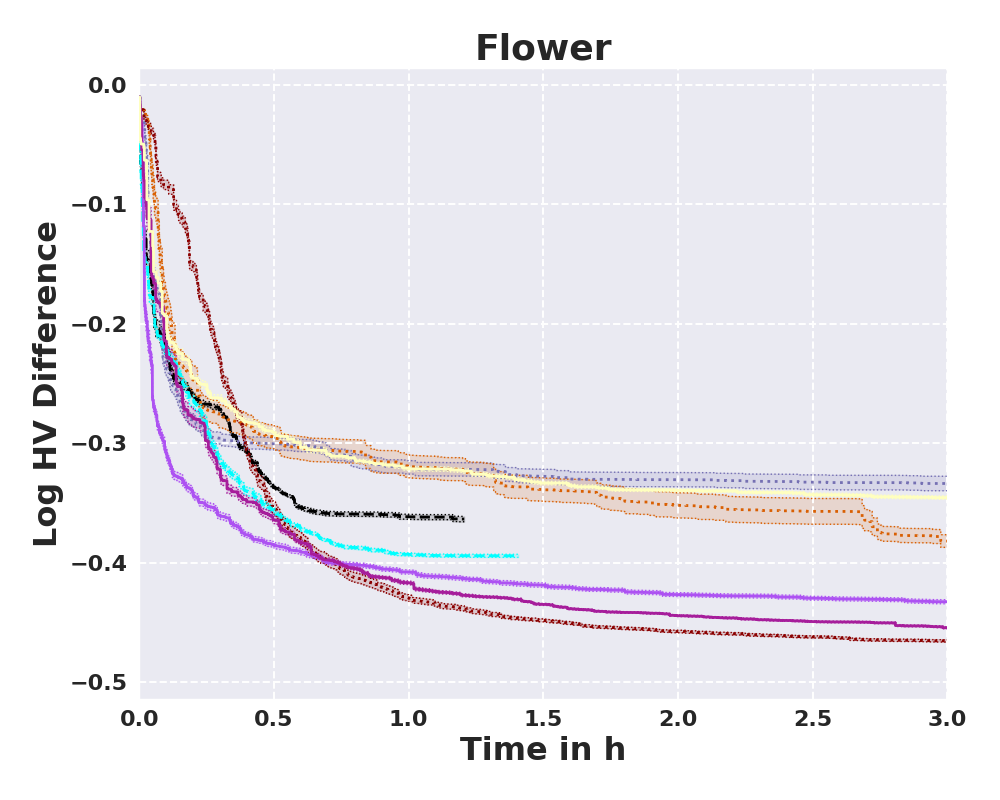}
       & \includegraphics[trim=0.8cm 1cm 0cm   0cm,clip,width=0.315\textwidth]{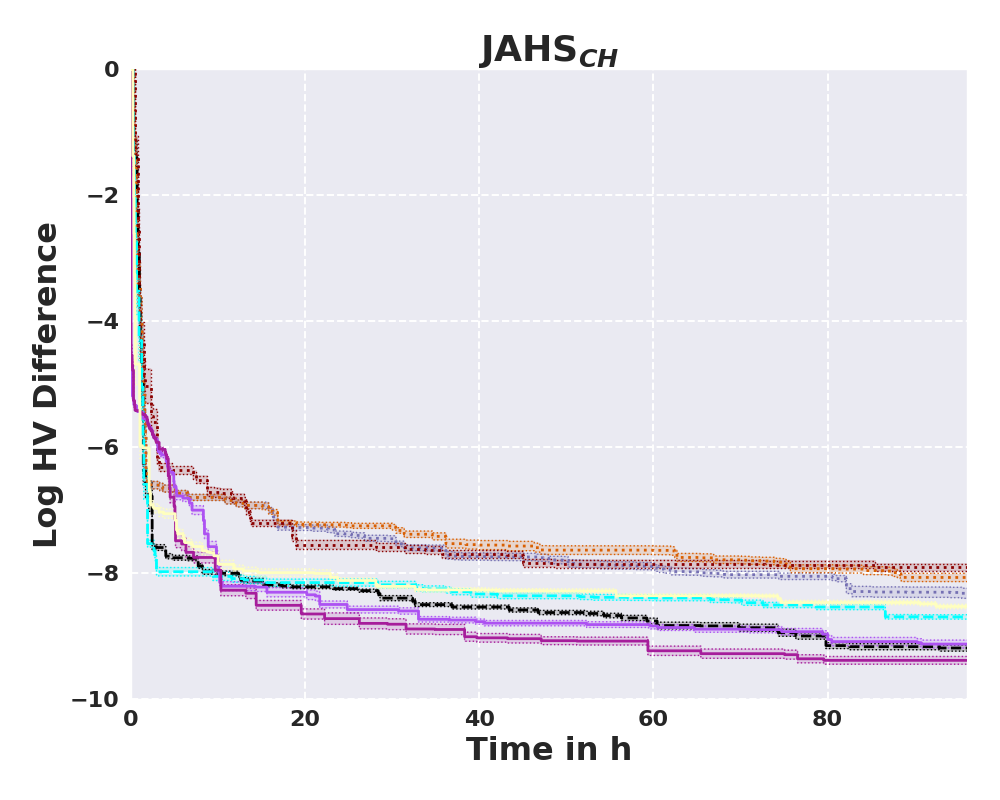} 
       & \includegraphics[trim=0.8cm 1cm 0cm  0cm,clip,width=0.33\textwidth]{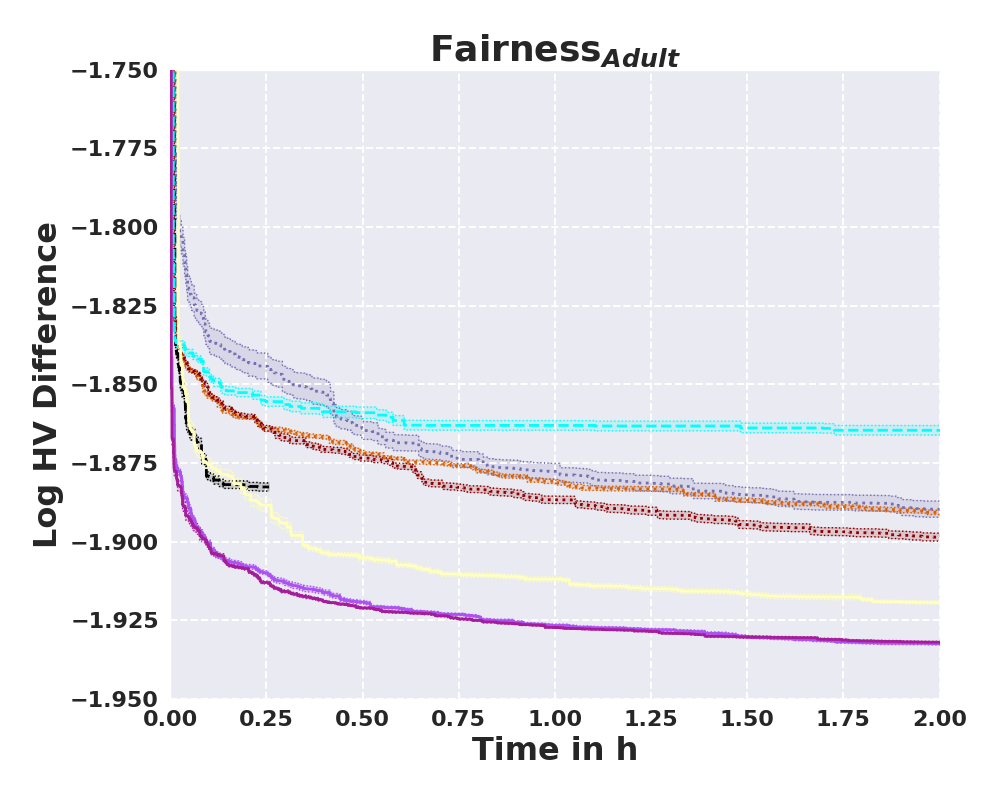}\\
  \end{tabular}
  \vspace{-0.3cm}
  \includegraphics[trim=0.9cm 0.5cm 0.5cm 0.5cm,clip, width=0.8\textwidth]  {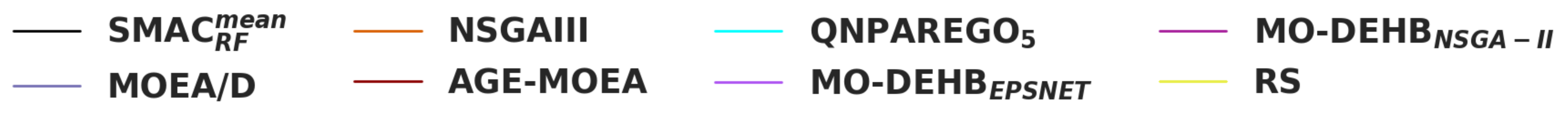}
  \end{center}
  \caption{Difference in dominated hypervolume of the Pareto front approximations for the different methods with respect to the combined front approximation over time on Imagenet and Cifar100 from NAS-Bench-201, Flowers, JAHS-Fashion and JAHS-Colorectal-Histology from Joint NAS \& HPO and Adult from algorithmic fairness}
  \label{fig:loghvdiff_comparison_all}
\end{figure*}

\begin{figure*}[ht]
  \begin{center}
  \begin{tabular}{l@{}l@{}l}
         \includegraphics[trim=0     1cm 0cm   0cm,clip,width=0.33\textwidth]{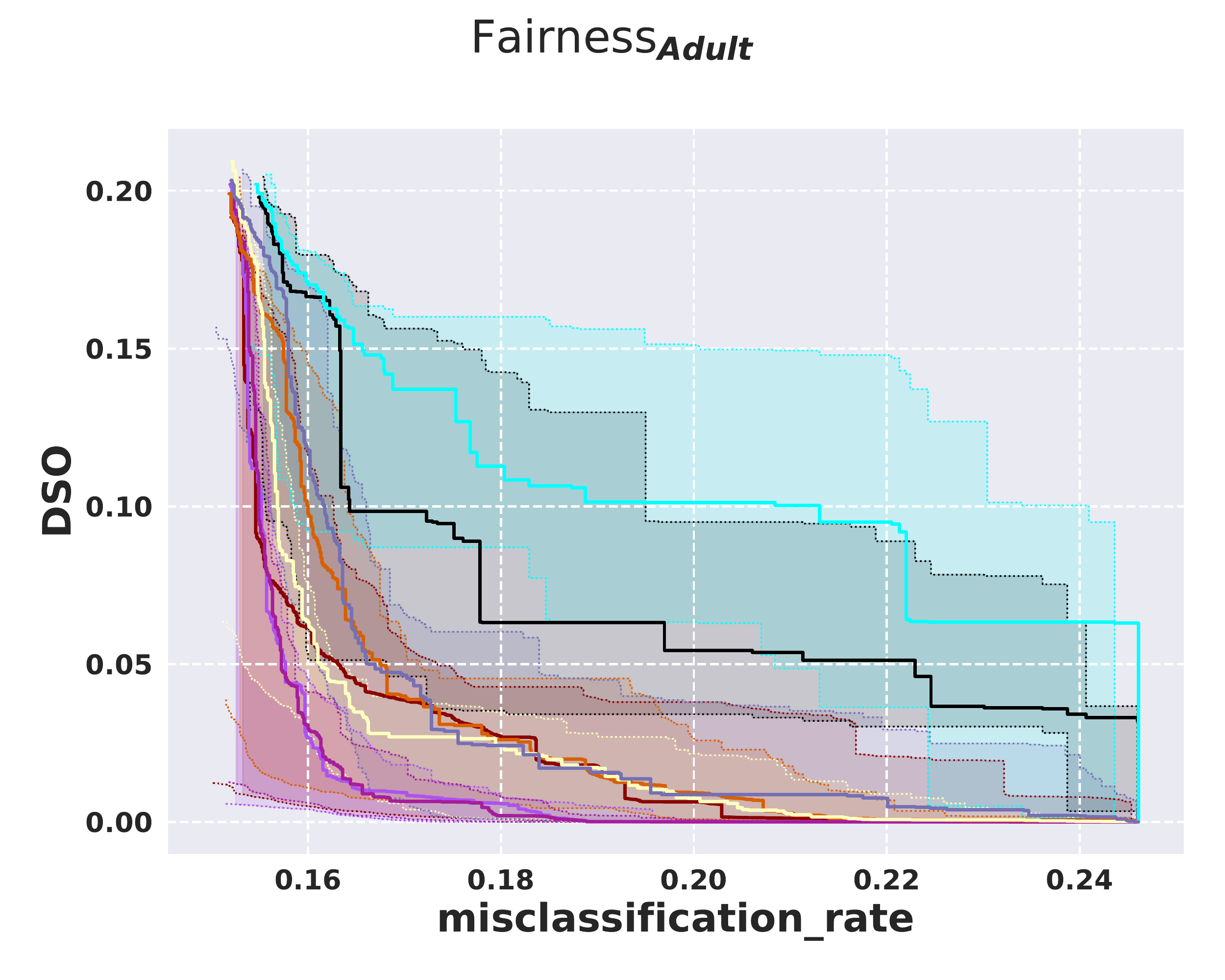}
       & \includegraphics[trim=0.8cm 1cm 0cm   0cm,clip,width=0.315\textwidth]{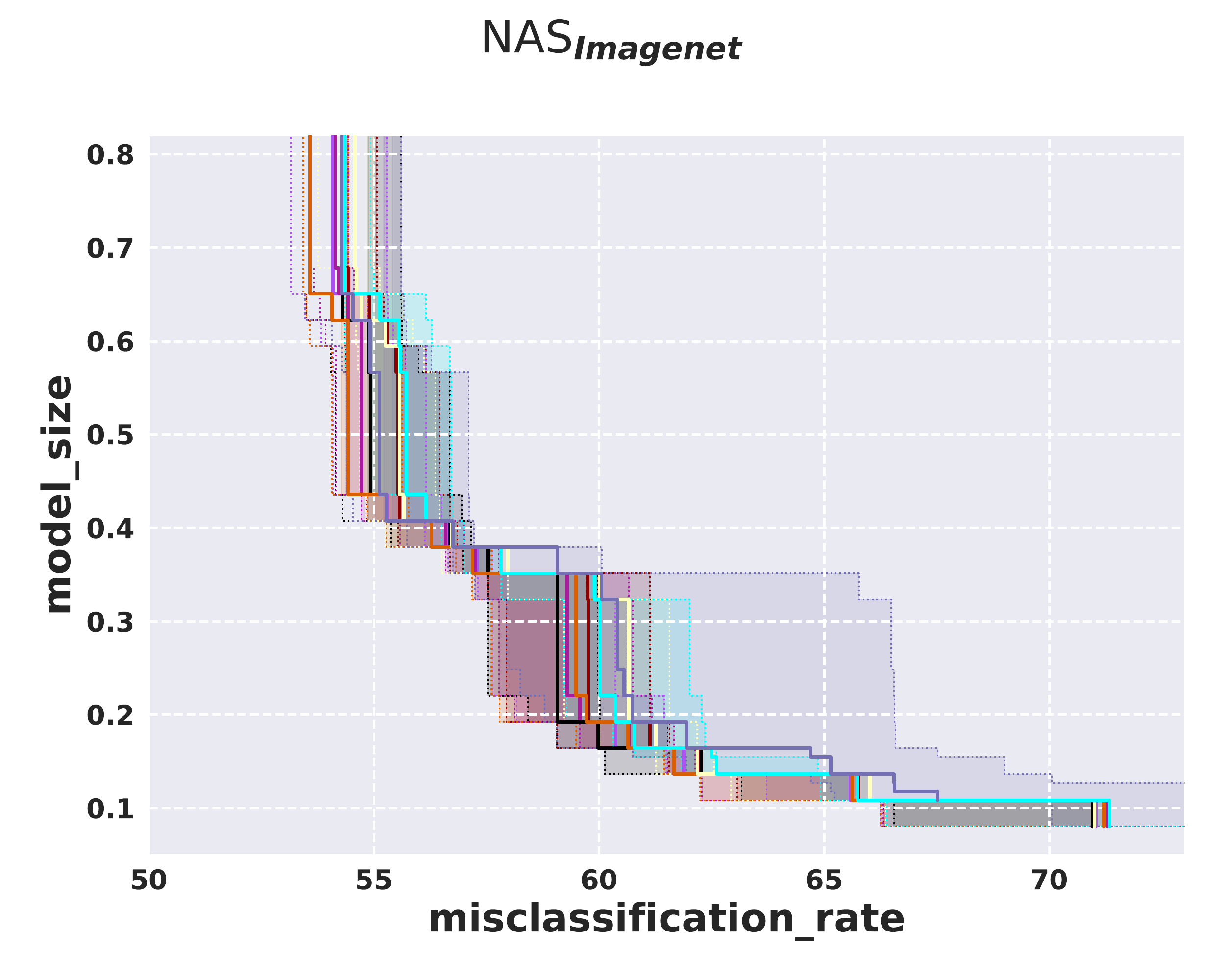}
       & \includegraphics[trim=0.8cm 1cm 0cm   0cm,clip,width=0.315\textwidth]{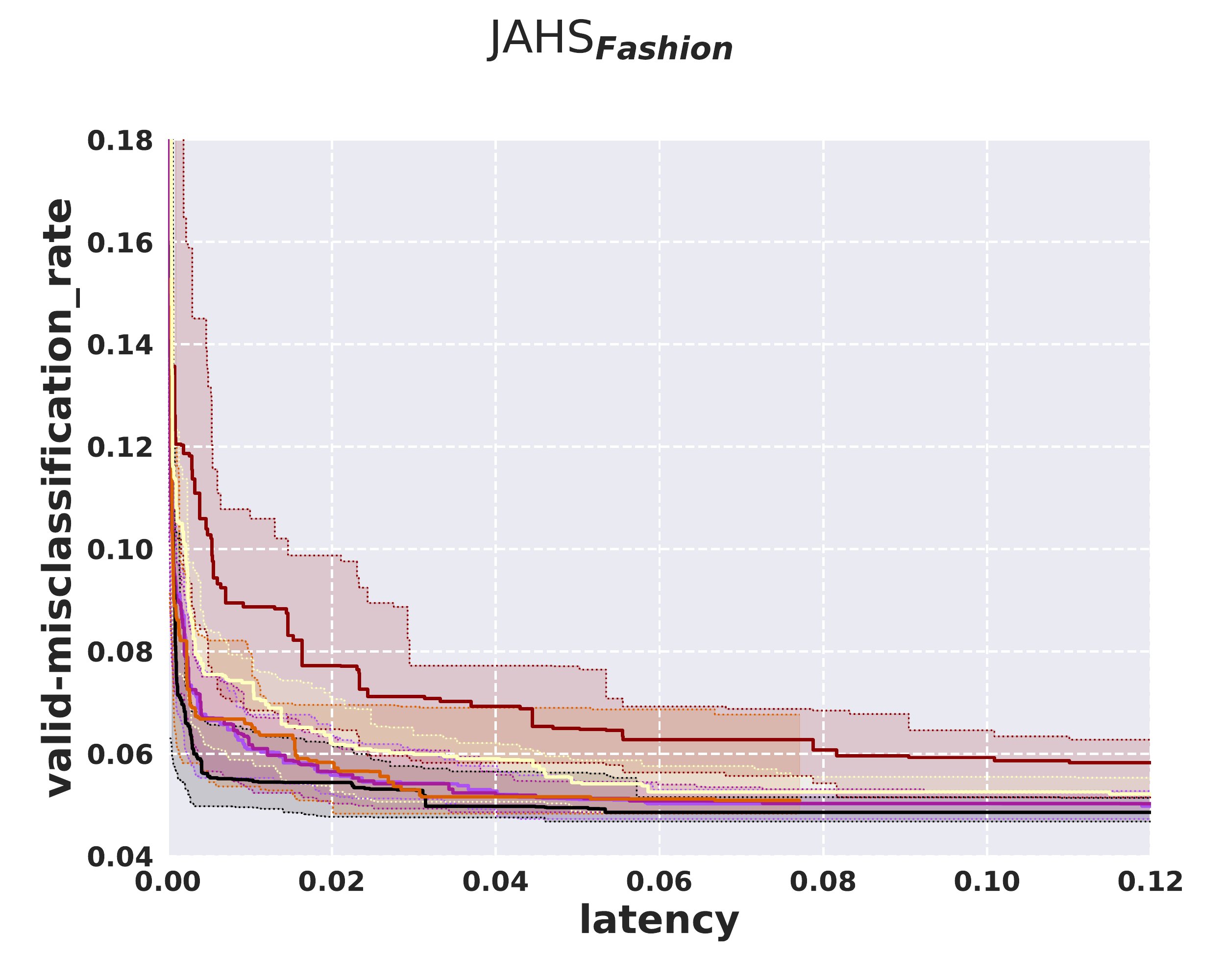} \\
         \includegraphics[trim=0     0.5cm 0cm 0cm,clip,width=0.33\textwidth]{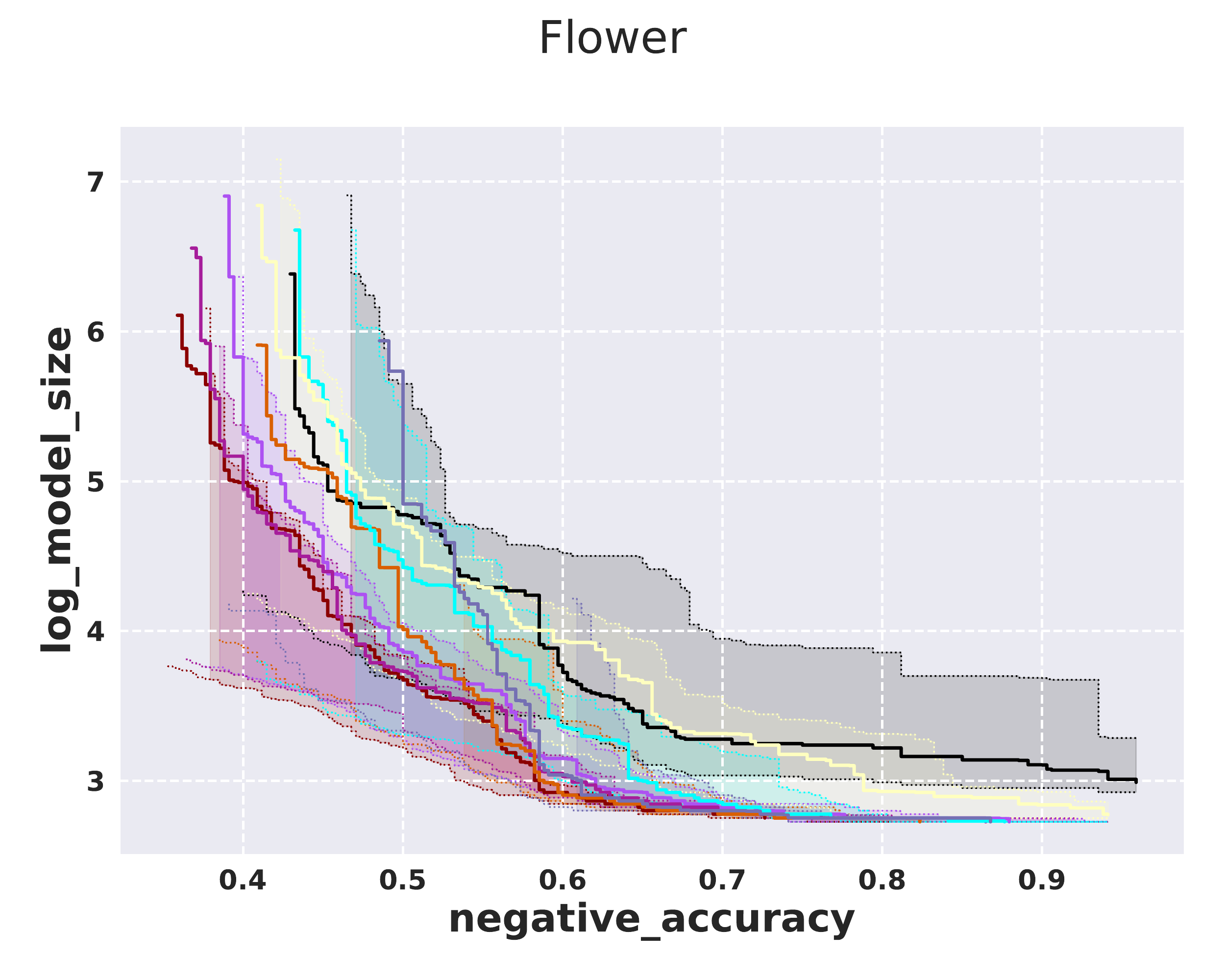}
       & \includegraphics[trim=0.8cm 0.5cm 0cm 0cm,clip,width=0.315\textwidth]{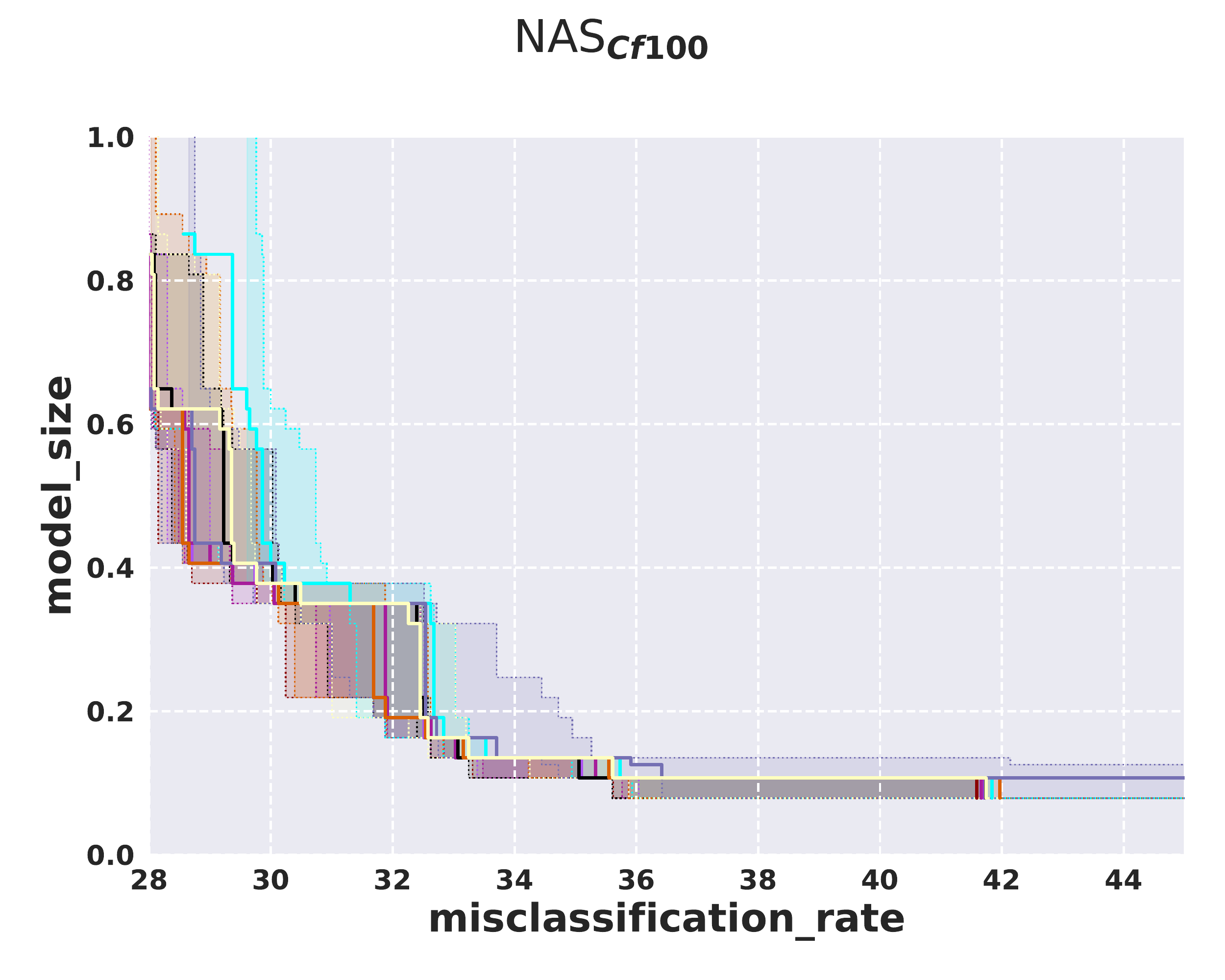}
       & \includegraphics[trim=0.8cm 0.5cm 0cm 0cm,clip,width=0.315\textwidth]{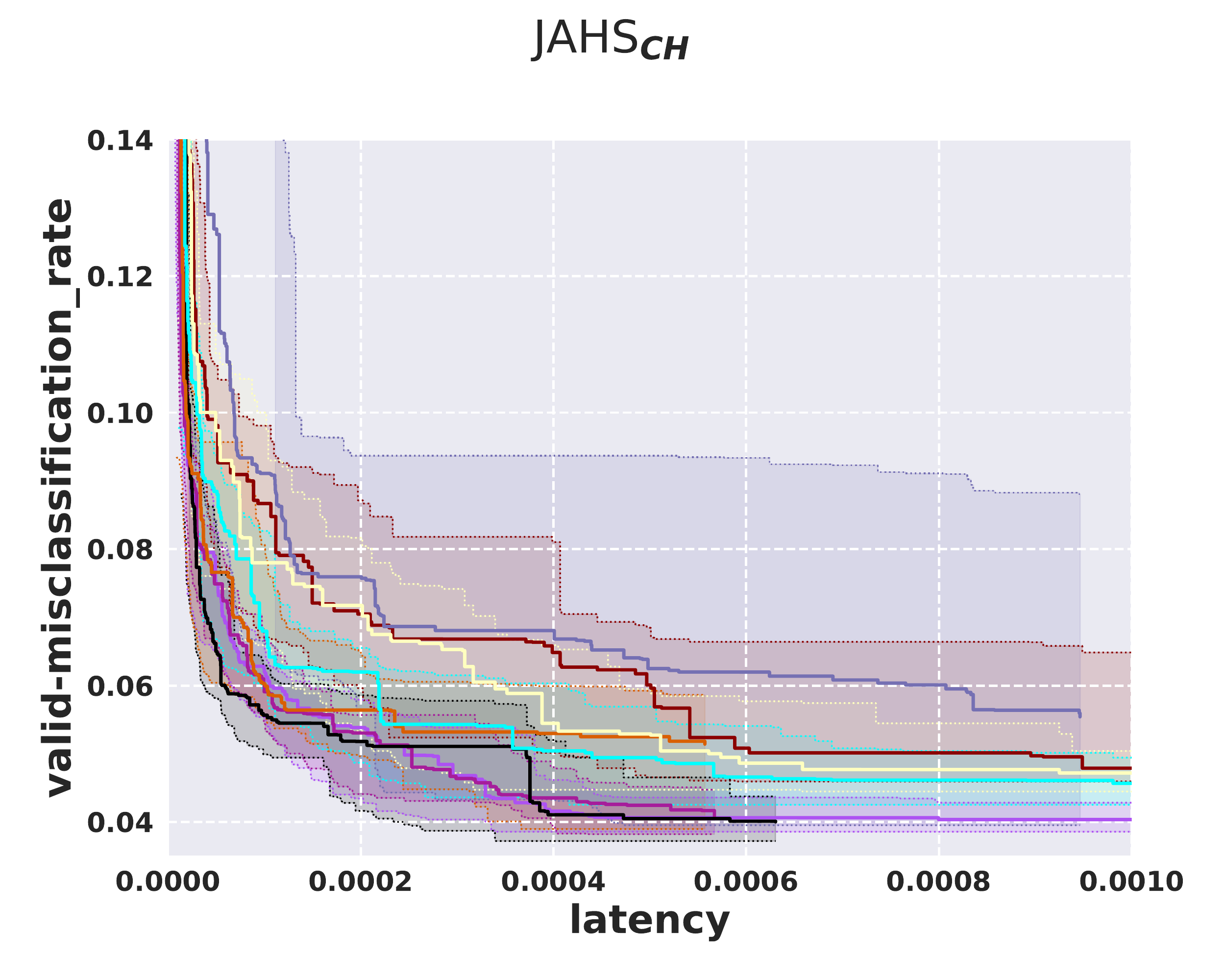} \\
  \end{tabular}
  \vspace{0cm}
  \includegraphics[trim=0.9cm 0.5cm 0.5cm 0cm,clip, width=0.8\textwidth]{plots/ranking_plots/Ranking_ijcai_all_ijcai_ff_100_legend.pdf}
  \end{center}
  \caption{We report summary-attainment-surfaces for on Imagenet and Cifar100 from NAS-Bench-201, Flowers, JAHS-Fashion and JAHS-Colorectal-Histology from Joint NAS \& HPO and Adult from algorithmic fairness. Upper and lower bound correspond to the first and ninth summary-attainment-surface.}
  \label{fig:attainment_surfaces}
\end{figure*}

\begin{figure}[htbp]
    \centering
    \includegraphics[width=0.98\columnwidth]{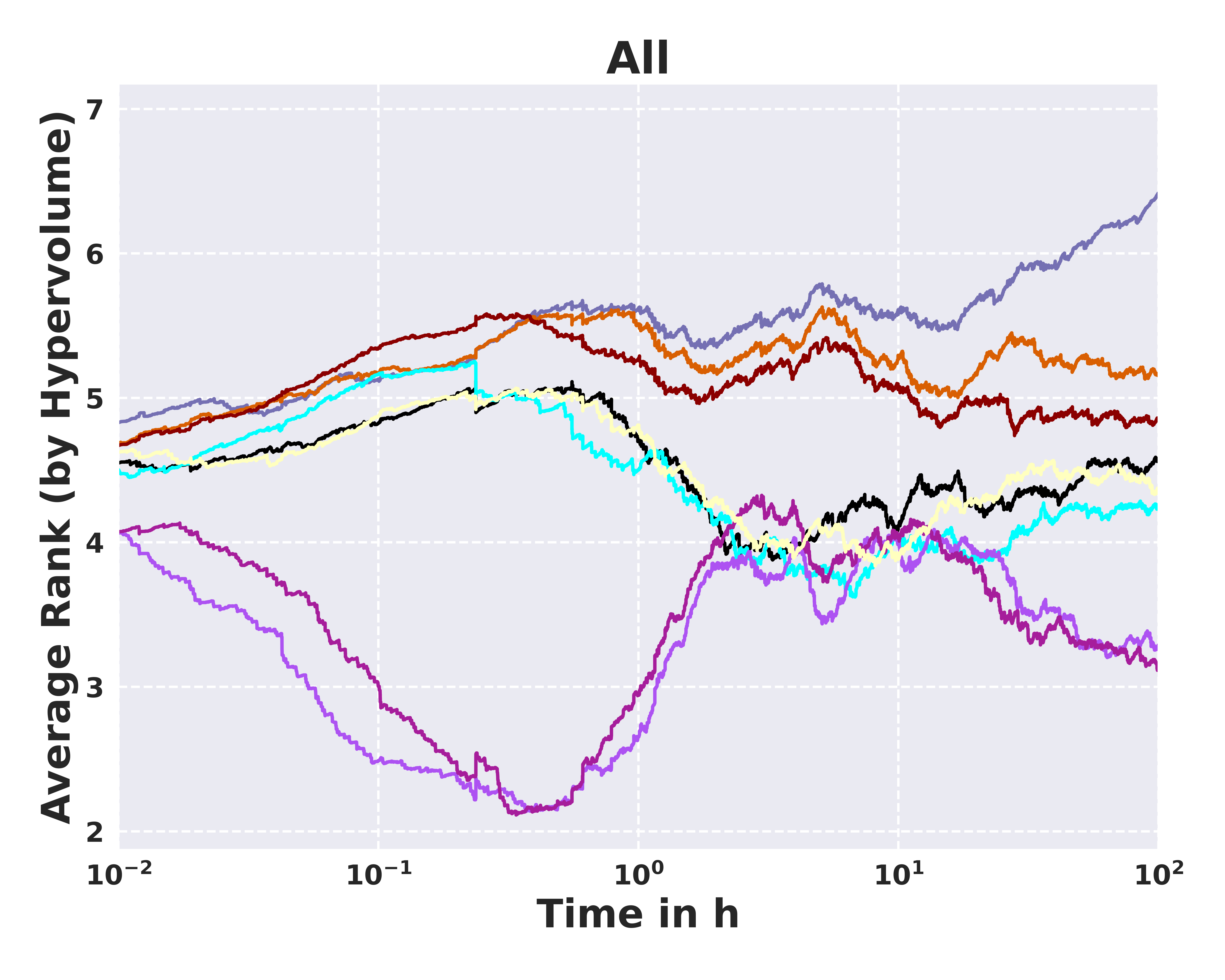}
  \includegraphics[trim=0.9cm 0.5cm 0.5cm 0cm,clip, width=\columnwidth]{plots/ranking_plots/Ranking_ijcai_all_ijcai_ff_100_legend.pdf}
    \caption{Average rank over time: We aggregate all 10 runs and compute the average rank over all benchmarks.}
    \label{fig:ranking_comparion}
\end{figure}

In this section, for space limitations, we present the results for 6 representative benchmarks: 2 from NAS (Imagenet and Cifar100 from NAS-Bench-201), 3 from joint NAS \& HPO (Flowers, JAHS-Fashion and JAHS-Colorectal-Histology) and algorithmic fairness on Adult dataset. The qualitatively similar plots for all the benchmarks can be found in Appendix C. 
Figure~\ref{fig:loghvdiff_comparison_all} presents the difference in dominated HV of the Pareto front approximations with respect to the combined front approximation over time on the 6 selected benchmarks for the different baseline methods. 
Overall, the \modehb{} variants clearly perform best. As seen in Figure~\ref{fig:loghvdiff_comparison_all}, both \modehb{} variants outperform all other methods in benchmarks such as JAHS-Fashion and fairness-Adult, showing exceptional performance throughout the optimization process. However, in benchmarks such as NAS-201-Cifar100 and JAHS-Colorectal-Histology, the \modehb{} variants performed poorly initially but later became competitive with other methods. Interestingly, \random{} performed well on the NAS-201-Imagenet benchmark, similar to \qnparego{}, and remained competitive till the end of optimization. Additionally, SMAC performed well during the first phase of exploration on this benchmark. The performance of the \modehb{} variants varied in some benchmarks, such as on the Flower dataset where \modehbepsnet{} performed well initially, but later \modehbnsgaii{} performed better, although it was outperformed by \agemoead{}.

Figure~\ref{fig:attainment_surfaces} visualizes a summary of attainment surfaces to assess the method's ability to cover the entire Pareto front~\cite{knowles2005summary}. We show the first, median and ninth attainment surfaces to visualize the distribution of Pareto fronts achieved by the methods to simplify visually inspecting the differences than plotting all 10 attainment surfaces. In Figure~\ref{fig:attainment_surfaces} for the Adult dataset, it can be seen that \modehb{} has a consistent performance as indicated by the small difference between the upper and lower bound. While SMAC performs better than the \modehb{} variants on the Fashion and Colorectal-Histology benchmarks from the JAHS-Bench, the \modehb{} variants still exhibit consistent and good performance. On the Imagenet benchmark, all methods perform similarly. This may be due to the high evaluation limits and the tabular nature of the benchmark.  



\subsection{Results Summary}
\label{subsec-resuls-summary}

We now compare \modehb{} variants and the other baseline methods across all the used benchmarks. We compute the average rank by hypervolume over time for each method in Figure~\ref{fig:ranking_comparion}. Both \modehb{} variants show exceptional performance and perform very well early on. Later during the optimization process and for a short period of time, \random{}, SMAC and \qnparego{} perform competitively. At the end, both \modehb{} variants show superior performance. We also report the mean and standard deviation for log HV difference for 10 repetitions across all the benchmarks for all the compared algorithms in Table~\ref{table:summary-mean-std}.   

\begin{table*}[!ht]
\centering
\setlength{\tabcolsep}{3pt}
\scriptsize 
\begin{tabular}{|c|c| c| c| c| c| c| c| c|} 
 \hline
 \  & RS & MOEA/D & AGE-MOEA & NSGAIII & SMAC & QNPAREGO & MO-DEHB$_{EPSNET}$ & MO-DEHB$_{NSGA-II}$ \\ [0.5ex]  
 \hline\hline
 NAS-101-CifarA & -2.897$\pm$0.042 & -2.684$\pm$0.161  & -2.901$\pm$0.033  & -2.885$\pm$0.045 & -2.858$\pm$0.028 & -2.889$\pm$0.038 &  -2.895$\pm$0.038 & \textbf{-2.916$\pm$0.0217}  \\ 
 \hline
 NAS101-CifarB & -2.887$\pm$0.0349 & -2.832$\pm$0.0515  & -2.888$\pm$0.0394  & -2.866$\pm$0.0297 & -2.890$\pm$0.0338 & -2.843$\pm$0.038 &  -2.893$\pm$0.025 & \textbf{-2.905$\pm$0.037}  \\ 
 \hline
 NAS101-CifarC & -2.896$\pm$0.040 & -2.752$\pm$0.178 & -2.876$\pm$0.041  & -2.870$\pm$0.038 & -2.856$\pm$0.038 & \textbf{-2.924$\pm$0.027} &  -2.889$\pm$0.037 & -2.894$\pm$0.031  \\ 
 \hline
 NAS1shot1-SS1 & -2.865$\pm$0.031 & -2.795$\pm$0.074  & -2.862$\pm$0.026  & -2.853$\pm$0.023 & -2.857$\pm$0.022 & -2.865$\pm$0.028 &  \textbf{-2.867$\pm$0.0182} & -2.851$\pm$0.029  \\ 
 \hline
 NAS1shot1-SS2 & -2.903$\pm$0.041 & -2.826$\pm$0.057  & -2.899$\pm$0.043  & -2.884$\pm$0.039 & -2.909$\pm$0.033 & \textbf{-2.913$\pm$0.026} &  -2.905$\pm$0.035 & -2.891$\pm$0.026  \\ 
 \hline
 NAS1shot1-imagenet & -0.487$\pm$0.024 & -0.487$\pm$0.024 & -0.487$\pm$0.024 & -0.487$\pm$0.024 & -0.484$\pm$0.021 & -0.475$\pm$0.016 & \textbf{-0.489$\pm$0.016} & -0.474$\pm$0.031  \\ 
 \hline
 NAS1shot1-cifar10 & -1.982$\pm$0.018 & -1.873$\pm$0.077 & -1.983$\pm$0.018 & -1.938$\pm$0.0402 & -1.986$\pm$0.027 & -1.986$\pm$0.0426 & \textbf{-1.998$\pm$0.043} & -1.99$\pm$0.025  \\ 
 \hline
 NAS1shot1-cifar100 & -1.066$\pm$0.013  & -1.051$\pm$0.030 & -1.066$\pm$0.013  & -1.070$\pm$0.024 & -1.073$\pm$0.017  & -1.069$\pm$0.013 & -1.069$\pm$0.027          & \textbf{-1.0914$\pm$0.0191}  \\ 
 \hline
 Fashion & -0.695$\pm$0.0945  & -0.469$\pm$0.392 & -0.286$\pm$0.120 & -0.538$\pm$0.325 & -0.740$\pm$0.056 & \textbf{-0.970$\pm$0.034} & \textbf{-0.919$\pm$0.043} & \textbf{-0.780$\pm$0.103}  \\ 
 \hline
 Flower & -1.066$\pm$0.013  & -1.051$\pm$0.030 & -1.066$\pm$0.013  & -1.070$\pm$0.024 & -1.073$\pm$0.017  & -1.069$\pm$0.013 & -1.069$\pm$0.027          & \textbf{-1.091$\pm$0.019}  \\ 
 \hline
JAHS-cifar10 & -7.201$\pm$0.636  & -6.059$\pm$0.628 & -5.665$\pm$0.633 & -6.096$\pm$0.661 & -7.807$\pm$0.601 & \textbf{-7.650$\pm$0.850} & \textbf{-7.891$\pm$0.649} & \textbf{-8.103$\pm$0.286}  \\ 
 \hline
JAHS-color-histology & -8.532$\pm$0.329 & -8.324$\pm$0.759 & \textbf{-7.910$\pm$0.533} & -8.071$\pm$0.618  & -9.191$\pm$0.468 & -8.697$\pm$0.334 & -9.128$\pm$0.596 & \textbf{-9.387$\pm$0.583}  \\ 
 \hline
 JAHS-fashion & -7.515$\pm$1.044 & -5.813$\pm$0.949 & -5.393$\pm$1.149 & -5.813$\pm$0.949 & -7.620$\pm$0.423 & -7.582$\pm$0.441 & -8.385$\pm$0.512 & \textbf{-8.430$\pm$0.342}  \\ 
 \hline
Adult & -1.919$\pm$0.004 & -1.889$\pm$0.024 & -1.898$\pm$0.012 & -1.890$\pm$0.006 & -1.882$\pm$0.012 & \textbf{-1.864$\pm$0.014} & \textbf{-1.932$\pm$0.004} & \textbf{-1.932$\pm$0.002}  \\ 
 \hline
\end{tabular}
\vspace*{0.2cm}
\caption{A summary of mean $\pm$ standard deviation for the log HV difference for $10$ receptions for all compared algorithms across all benchmarks}
\label{table:summary-mean-std}
\end{table*}

\section{Conclusion}
\label{sec-conclusion}
In this paper, we extended \dehb{}, a very effective evolutionary-based hyperband, to multi-objective optimization in the context of HPO. We use techniques that optimize the Pareto geometry based on non-dominating sorting and propose two variants: (1) \modehbnsgaii{} and (2) \modehbepsnet{}. We evaluate the performance of \modehb{} using comprehensive benchmark suites consisting of diverse and challenging MO problems from three benchmark families: NAS, NAS+HPO and algorithmic fairness. We demonstrated that \modehb{} outperforms state-of-the-art MO optimizers on most of the benchmark problems. These results suggest that \modehb{} is a promising new MO optimizer that is well-suited to a wide range of real-world optimization problems. For future work, we aim to propose an asynchronous version of \modehb{} that takes advantage of parallel resources as an improved version of the original parallel settings in \dehb{}. Our reference implementation of \modehb{} is available at \url{https://anonymous.4open.science/r/MODEHB-F38E}.  



\section*{Ethical Statement}

There are no ethical issues.


\newpage
\bibliographystyle{named}
\bibliography{bib/shortstrings,bib/lib,bib/local,bib/shortproc}

\end{document}


\maketitle
\appendix




\section{More Details on Experiments}
\label{appendix:sec-experiments}

\subsection{Baseline Methods}
\label{appendix:subsec-benchmarks}
In our experiments, we use default settings to run the baselines methods as it is reported in their papers or implementations such as  \cite{blank2020pymoo}, \cite{balandat2020botorch}.

\textbf{Random Search (RS)}
In each generation, random architectures are sampled from the configuration space using a uniform distribution.

\textbf{\qnparego{}}
We use the implementation from \cite{balandat2020botorch}. We use 20 initial samples and a batch size of 5

\textbf{\smacrfmean{}}
We use the implementation from
\href{https://github.com/automl/SMAC3}
{https://github.com/automl/SMAC3}
we use HyperparameterOptimizationFacade with MeanAggregationStrategy  as the multi-objective algorithm

\textbf{\nsgaiii{}}
We use the implementation for \nsgaiii{} from pymoo repository\cite{blank2020pymoo}. we use 10 partitions to generate reference direction. 

\textbf{\agemoead{}}
We used the implementation for \agemoead{} from pymoo repository\cite{blank2020pymoo}. We use a population of size 100

\textbf{\moead{}}
We used the implementation for \moead{} from pymoo repository\cite{blank2020pymoo}. we use 10 partitions to generate reference direction with the default setting of number of neighbors as 20. We use auto decomposition and a default setting for probability of neighbor mating as 0.9 

\subsection{Benchmarks}
\label{appendix:subsec-benchmarks}
We collect benchmarks for multi-objective (MO) that optimize interesting objectives from three diverse domains: \nas{} (NAS), joint NAS and hyperparameter optimization (joint NAS \& HPO), and algorithmic fairness. We build our collection of benchmarks on \hpobench{} library~\cite{eggensperger-neuripsdbt21a}. For NAS family, we conduct experiments on NAS-Bench-101~\cite{ying2019bench}, NAS-Bench-1shot1~\cite{zela2020bench} and NAS-Bench-201~\cite{dong2020bench}, which are 9 tabular benchmarks.
The joint NAS \& HPO family involving tuning Convolutional Neural Networks (CNNs) the Oxford-Flowers dataset~\cite{Nilsback08} and Fashion-MNIST~\cite{xiao2017fashion}, and also three surrogate benchmarks~\cite{zela-iclr22a} from the recently introduced JAHS-Bench-201 suite~\cite{bansal2022jahs}. For algorithmic fairness, we have a fair model adopted from~\cite{Schmucker-arxiv21} on Adult dataset~\cite{kohavi1996scaling}. 
In Table \ref{appendix-tab:benchmarks-summary}, we provide a summary for all the benchmarks with details on search space and its type, optimized objectives and fidelity. 

\begin{table*}[tbp]
\centering
\small
\begin{tabular}{@{\hskip 0mm}
l@{\hskip 0mm}c
c@{\hskip 1mm}c@{\hskip 1mm}c@{\hskip 1mm}c
c@{\hskip 1mm}c@{\hskip 1mm}c@{\hskip 1mm}c
c@{\hskip 1mm}c
@{\hskip 0mm}}
\toprule
               Family & \#benchs & \#cont(log) & \#int(log) & \#cat & \#ord & fidelity & type & objectives & opt. budget & \#confs & Ref. \\
\midrule
\multirow{3}{*}{NAS101} & \multirow{3}{*}{3} & 0 &   0 &  26 &   0 &  \multirow{3}{*}{epochs} & \multirow{3}{*}{Tabular} & Accuracy & $10^7$sec & \multirow{3}{*}{$423$k} & \multirow{3}{*}{\cite{ying-icml19a}} \\
{} & {} & 0 &   0 &  14 &   0 & {}  & {}  & \multirow{2}{*}{Modelsize}  & 428 TAE & {} & {}  \\
{} & {} & 21 &   1 &   5 &   0 & {}  & {}  & {}  & 435 TAE  & {} & {}  \\
\midrule
\multirow{2}{*}{NAS201}     & \multirow{2}{*}{3}                 &   \multirow{2}{*}{0} &   \multirow{2}{*}{0} &   \multirow{2}{*}{6} &   \multirow{2}{*}{0}&  \multirow{2}{*}{epochs} & \multirow{2}{*}{Tabular} & Accuracy & $10^7$sec & \multirow{2}{*}{$15\ 625$} & \multirow{2}{*}{\cite{dong-iclr20a}} \\
                            &                    &                          &                       &                        &                        &                            &                            &  Modelsize     &  216 TAE    &   &  \\
\midrule
\multirow{3}{*}{NAS1shot1} & \multirow{3}{*}{3} & \multirow{3}{*}{0} & \multirow{3}{*}{0} & 9 & \multirow{3}{*}{0} & \multirow{3}{*}{epochs} & \multirow{3}{*}{Tabular} & Accuracy & $10^7$sec & $6\ 240$ & \multirow{3}{*}{\cite{zela-iclr20b}} \\
{} &  &  &  & 9  & {} & {} & {} & \multirow{2}{*}{Model size} & 260 TAE & $29\ 160$ & {} \\
{} &  &  &  & 11 & {} & {} & {} & {} & 285 TAE & $363\ 648$ & {} \\

\midrule
\multirow{2}{*}{Joint Nas\&HPO} & \multirow{2}{*}{2}  & \multirow{2}{*}{1(1)}   &  \multirow{2}{*}{9(7)}   &   \multirow{2}{*}{3}   &   \multirow{2}{*}{0}   &   \multirow{2}{*}{epochs}  & \multirow{2}{*}{raw}      &  Accuracy    &   86400 sec  & \multirow{2}{*}{-} & \multirow{2}{*}{\cite{izquierdo2021bag}} \\
                            &                    &                          &                       &                        &                        &                            &                            &  Log Modelsize     &   309 TAE   &   & {} \\
\midrule
\multirow{2}{*}{JAHS-Bench-201}  & \multirow{2}{*}{3} & \multirow{2}{*}{2(2)}    &  \multirow{2}{*}{0}   &   \multirow{2}{*}{9}   &   \multirow{2}{*}{3}   &   \multirow{2}{*}{epochs}  & \multirow{2}{*}{surrogate} &  Accuracy    &   $10^7$ sec & \multirow{2}{*}{$200$k} & \multirow{2}{*}{\cite{bansal2022jahs}} \\
                            &                    &                          &                       &                        &                        &                            &                            &  Latency     &   320 TAE   &   &  \\
\midrule
\multirow{2}{*}{Fairness$_{Adult}$} & \multirow{2}{*}{1}  & \multirow{2}{*}{5(5)}   &  \multirow{2}{*}{5(4)}   &   \multirow{2}{*}{0}   &   \multirow{2}{*}{0}   &   \multirow{2}{*}{epochs}  & \multirow{2}{*}{raw}      &  Accuracy    &   86400 sec  & \multirow{2}{*}{-} & \multirow{2}{*}{\cite{schmucker2021multi}} \\
                            &                    &                          &                       &                        &                        &                            &                            &  DSO     &  273 TAE    &   &  \\

\bottomrule
\end{tabular}
\caption{Overview of used benchmark. We report the number of benchmarks per family (\emph{\#benchs}), the number of continuous (\emph{\#cont}), integer (\emph{\#int}), categorical (\emph{\#cat}), ordinal (\emph{\#ord}) hyperparameters and if they are on a log scale. We also report benchmark type, optimization objectives and budgets. We set a upper limit per benchmark of Target Algorithm Executions (TAE) depending on the search space ($20 + 80 * \sqrt{|\text{Search Space}|}$) 
}
\label{appendix-tab:benchmarks-summary}
\end{table*}

\subsection{Results for Neural Architecture Search}
\label{appendix:subsec-results-NAS} 
\begin{figure}[H]
    \centering
    \includegraphics[width=0.80\columnwidth, trim=0.9cm 1.65cm 0.9cm 0.9cm,clip]{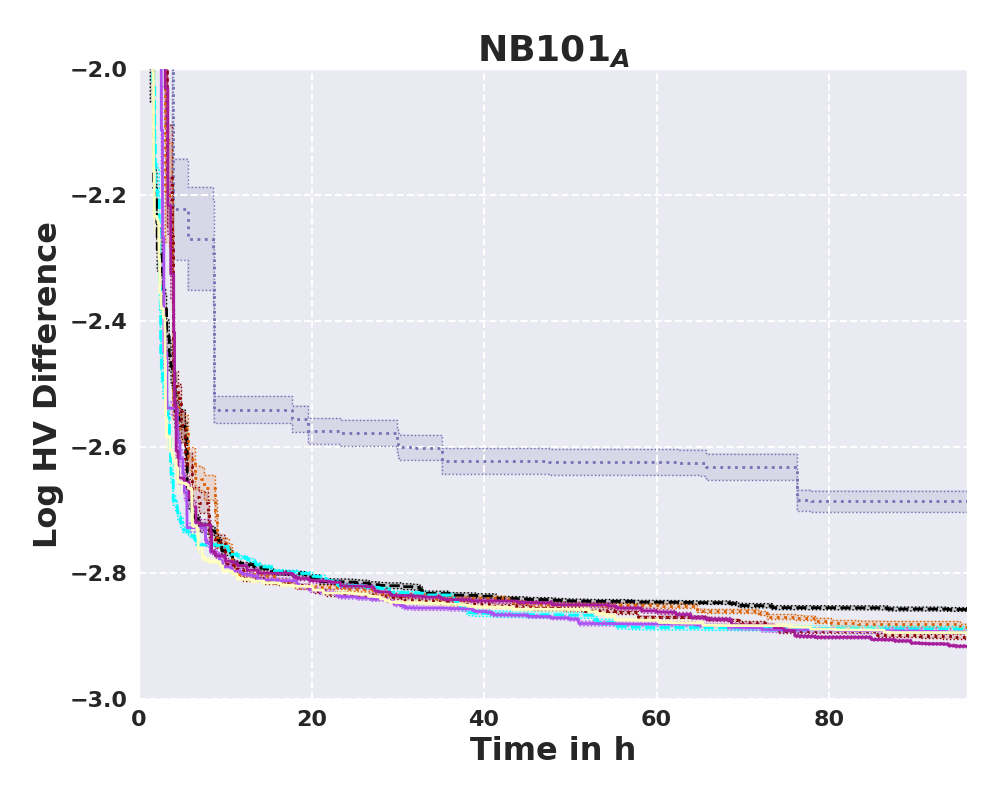}
    \includegraphics[width=0.80\columnwidth, trim=0.9cm 1.65cm 0.9cm 0.9cm,clip]{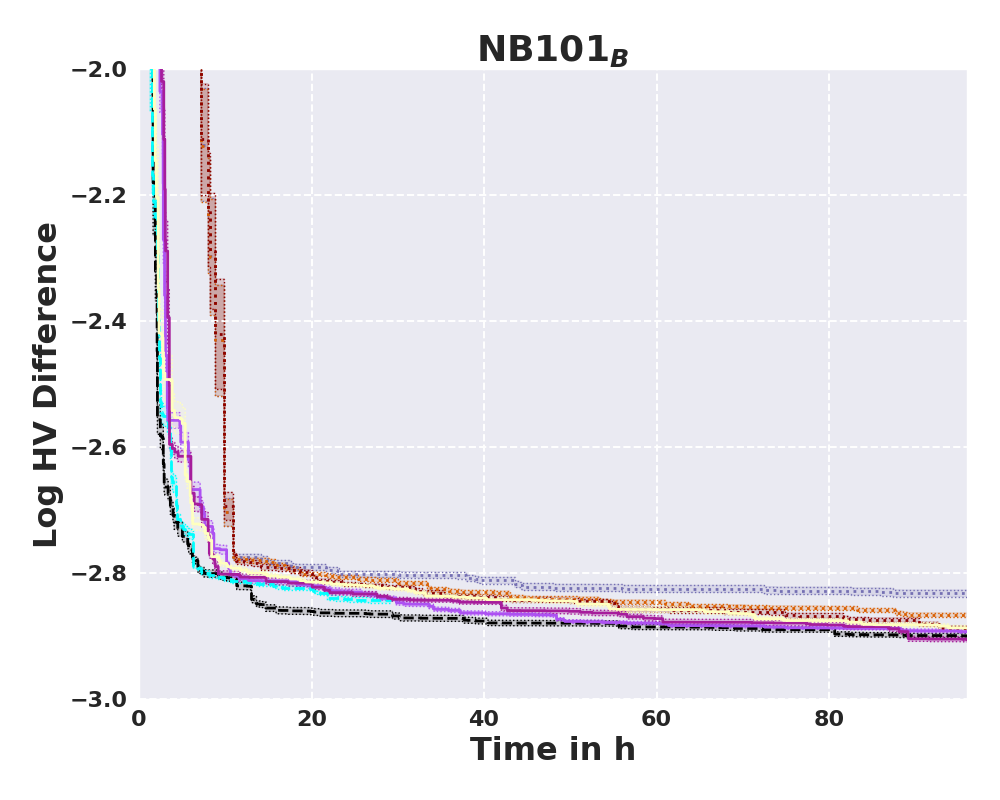}
    \includegraphics[width=0.80\columnwidth, trim=0.9cm 0.9cm 0.9cm 0.9cm,clip]{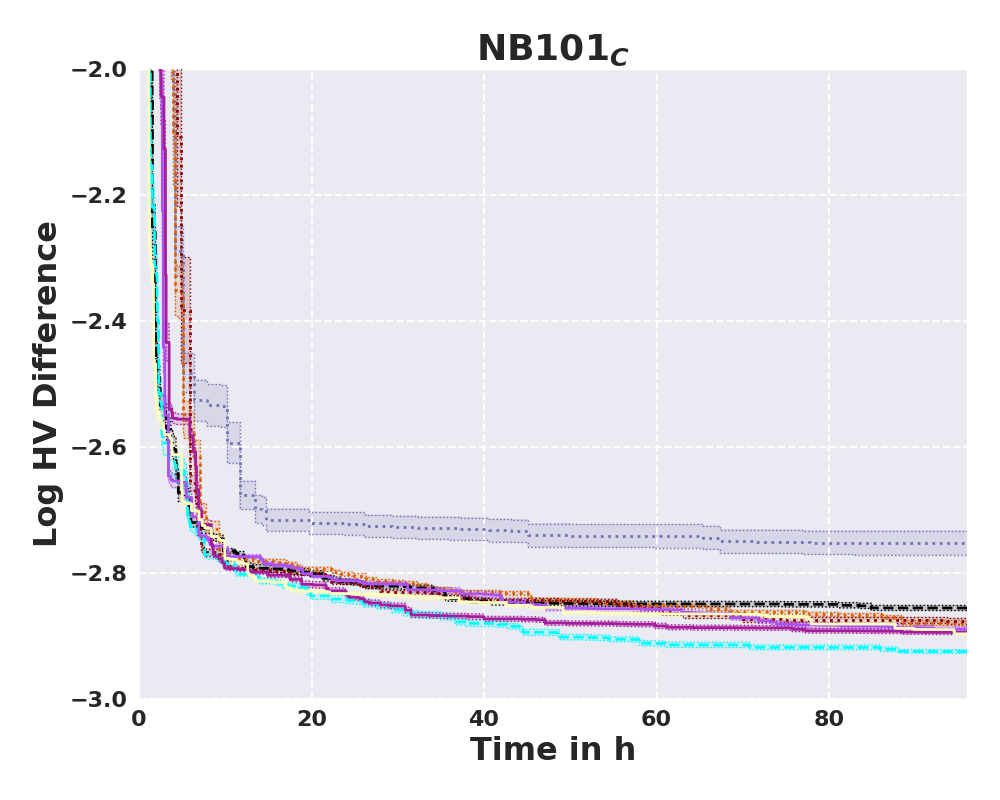}
    \caption{Log HV Differences between empirical best and trajectory on NAS-Bench 101.}
    \label{fig:loghv nas101}
\end{figure}

\begin{figure}[H]
    \centering
    \includegraphics[width=0.80\columnwidth, trim=0.9cm 1.65cm 0.9cm 0.9cm,clip]{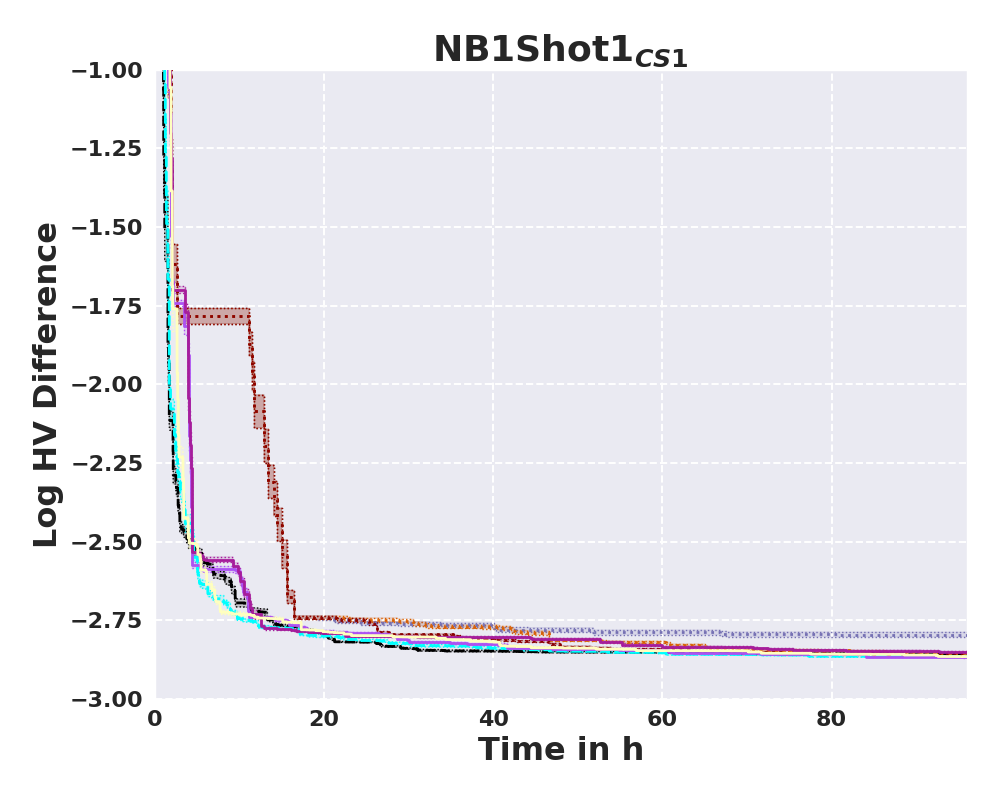}
    \includegraphics[width=0.80\columnwidth, trim=0.9cm 0.9cm 0.9cm 0.9cm,clip]{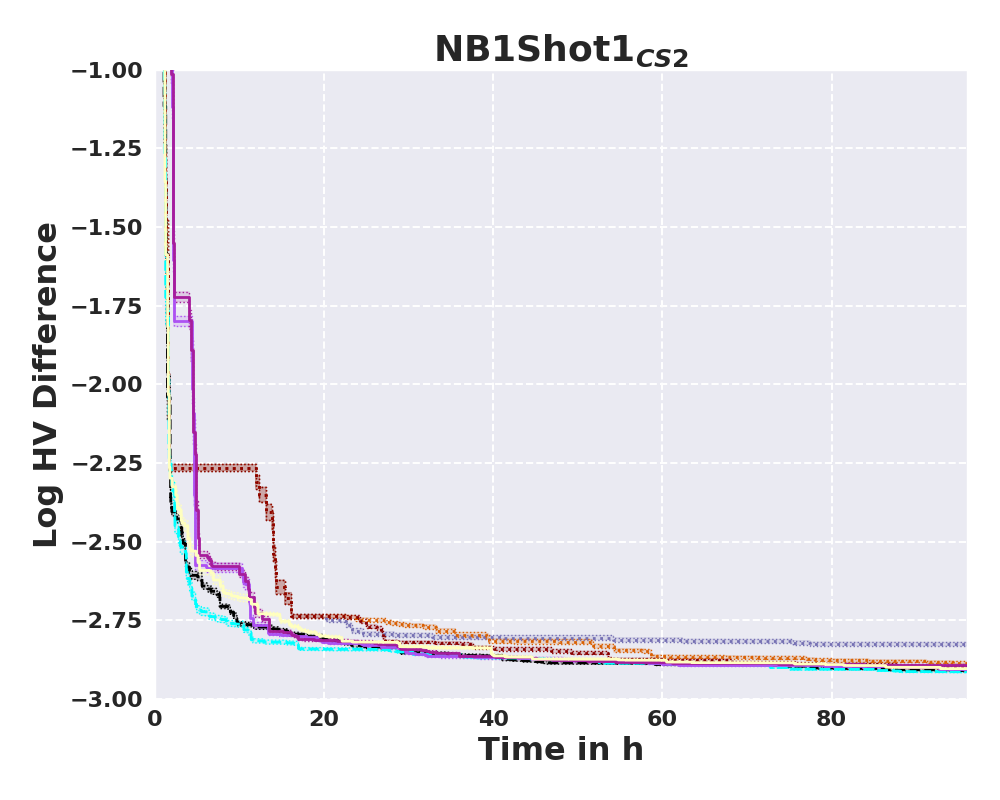}
    \caption{Log HV Differences between empirical best and trajectory on NAS-Bench 1shot1}
    \label{fig:loghv nas1shot1}
\end{figure}

\begin{figure}[H]
    \centering
    \includegraphics[width=0.80\columnwidth, trim=0.9cm 1.65cm 0.9cm 0.9cm,clip]{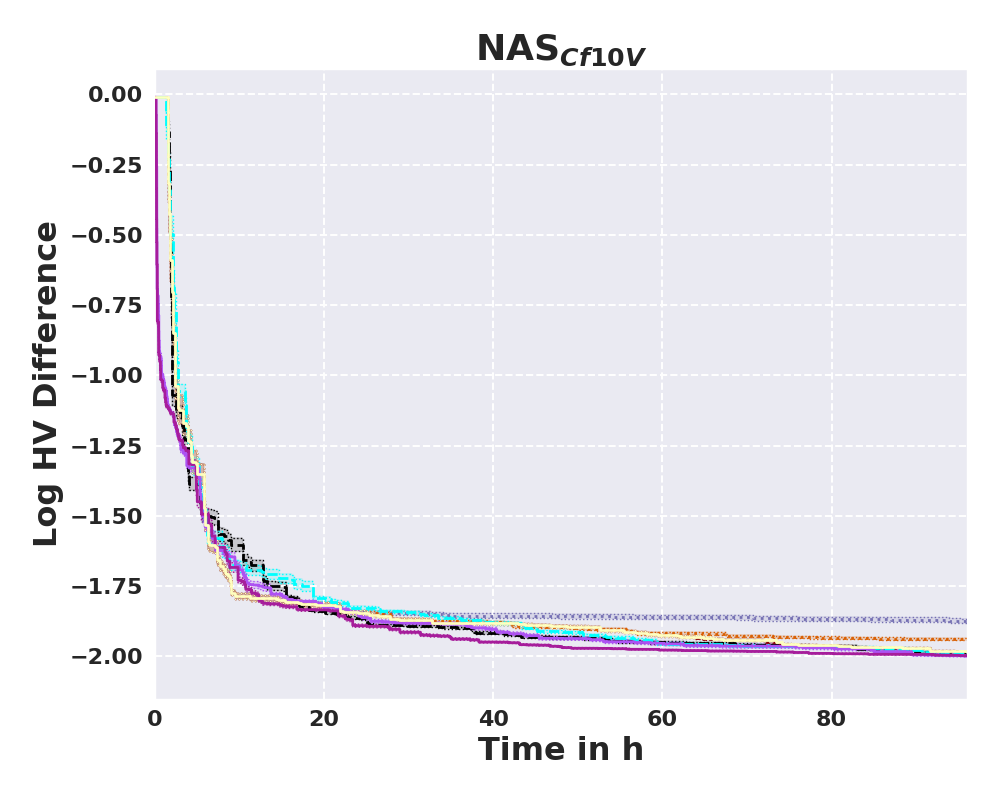}
    \includegraphics[width=0.80\columnwidth, trim=0.9cm 1.65cm 0.9cm 0.9cm,clip]{plots/log_hv_plots/LogHVDiff_mo_nas_201_cifar100_walltime_legend_disabled.png}
    \includegraphics[width=0.80\columnwidth, trim=0.9cm 0.9cm 0.9cm 0.9cm,clip]{plots/log_hv_plots/LogHVDiff_mo_nas_201_imagenet_valid_walltime_legend_disabled.png}
    \caption{Log HV Differences between empirical best and trajectory on NAS-Bench 201}
    \label{fig:loghv nas201}
\end{figure}

In Figure~\ref{fig:loghv nas101}, the performance of all baseline algorithms is evaluated on NAS-Bench-101. We observe that all baseline algorithms perform similarly except for \moead{}. \modehbnsgaii{} slightly outperforms the rest on NAS-Bench-101-A and NAS-Bench-101-B, while \qnparego{} demonstrates the best overall performance on NAS-Bench-101-C.
Figure~\ref{fig:loghv nas1shot1} presents the results for NAS-Bench-1Shot1\footnote{Due to minor integration issues, the observation for 1Shot\_CS\_3 is currently unavailable. However, it will be provided in the near future}. We observe that that all baseline algorithms, except \moead{}, converge with a similar performance . 
Figure~\ref{fig:loghv nas201} presents the results on NAS-201. For Imagenet benchmark, we see that \random{} serves as a strong baseline. Also, however the \modehb{} variants perform poorly initially for a short period of time, later they converge to a similar performance compared to other baselines. Furthermore, \smacrfmean{} demonstrates a strong performance on all benchmarks, although it is slightly outperformed by \modehb{} on NAS-201-Cifar100. 

\subsection{Results for Joint NAS \& HPO}
\label{appendix:subsec-results-NAS+HPO} 
\begin{figure}[H]
    \centering
    \includegraphics[width=0.80\columnwidth, trim=0.9cm 1.65cm 0.9cm 0.9cm,clip]{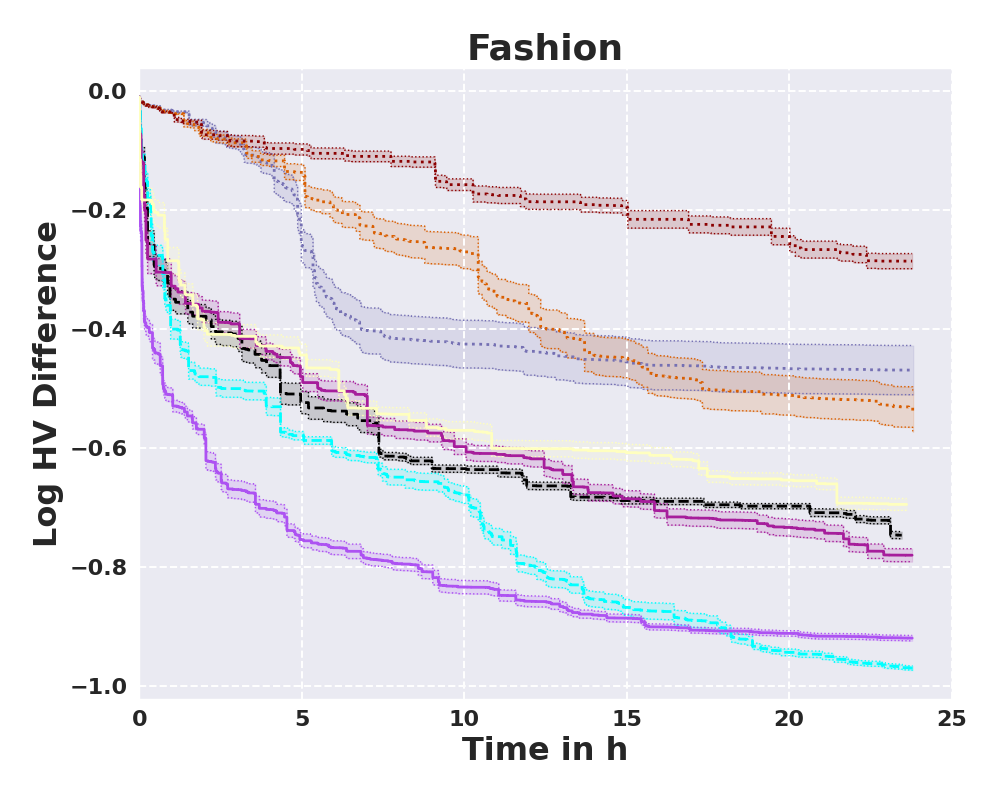}
    \includegraphics[width=0.80\columnwidth, trim=0.9cm 0.9cm 0.9cm 0.9cm,clip]{plots/log_hv_plots/LogHVDiff_mo_cnn_flower_walltime_legend_disabled.png}
    \caption{Log HV Differences between empirical best and trajectory on joint NAS \& HPO for CNN Fashion and Flower datasets}
    \label{fig:loghv mocnn}
\end{figure}

\begin{figure}[H]
    \centering
    \includegraphics[width=0.80\columnwidth, trim=0.9cm 1.65cm 0.9cm 0.9cm,clip]{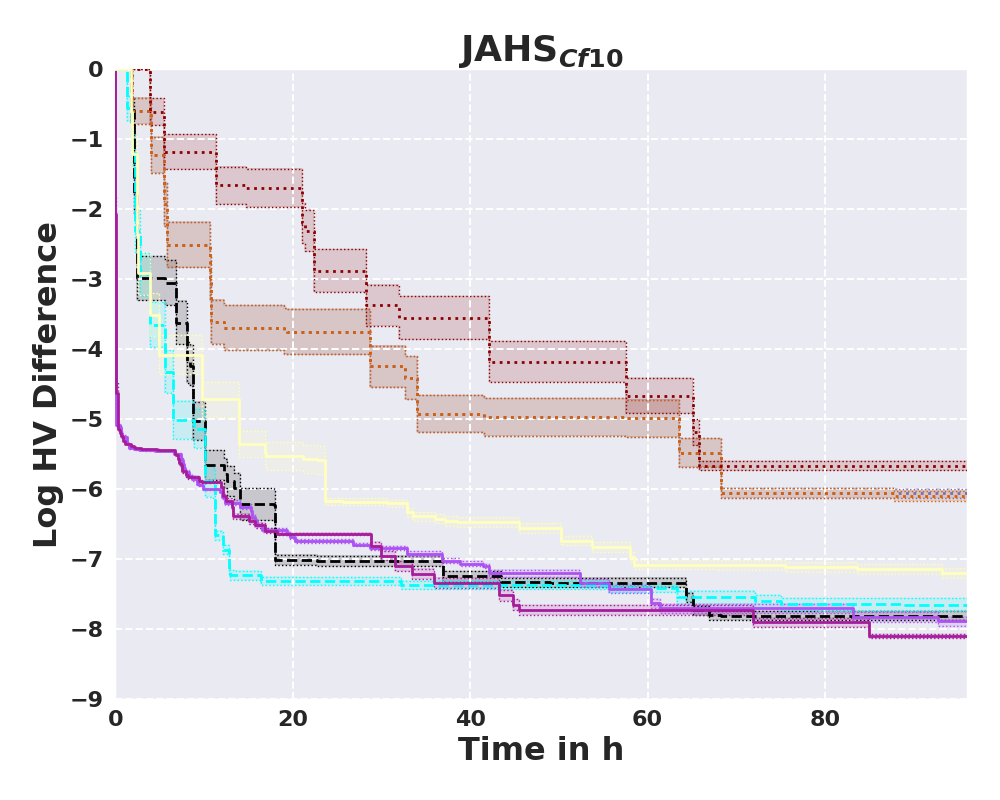}
    \includegraphics[width=0.80\columnwidth, trim=0.9cm 1.65cm 0.9cm 0.9cm,clip]{plots/log_hv_plots/LogHVDiff_mo_jahs_fashion_walltime_legend_disabled.png}
    \includegraphics[width=0.80\columnwidth, trim=0.9cm 0.9cm 0.9cm 0.9cm,clip]{plots/log_hv_plots/LogHVDiff_mo_jahs_color_histology_walltime_legend_disabled.png}
    \caption{Log HV Differences between empirical best and trajectory on JAHS-Bench-201 Benchmark}
    \label{fig:loghv jahs}
\end{figure}

 Figure~\ref{fig:loghv mocnn} presents the results for Fashion and Flower datasets. We observe that on Flower benchmark, \modehbepsnet{} performed well initially but it is outperformed by \modehbnsgaii{} later, with \agemoead{} showing the final best performance. On the Fashion dataset, we see that while  \modehbepsnet{} consistently outperforms all other baseline methods, it is outperformed by \qnparego{} in the end of optimization. In Figure~\ref{fig:loghv jahs} we show the results for JAHS-Bench-201 suite. We observe that \modehbnsgaii{} shows the final best performance on all three benchmarks while \modehbepsnet{} performs competitively. Additionally, we observe that \smacrfmean{} exhibits competitive performance on JAHS-Cifar10 and JAHS-Colorectal-Histology.

\subsection{Summary of Results}
\begin{figure*}[ht]
  \begin{center}
  \begin{tabular}{l@{}l@{}l}
       \includegraphics[trim=0.8cm 1cm 0cm   0cm,clip,width=0.315\textwidth]{plots/attainment_surfaces/mo_nas_201_imagenet_valid_attainment_surfaces_CI__3C.pdf}
       & \includegraphics[trim=0.8cm 0.5cm 0cm 0cm,clip,width=0.315\textwidth]{plots/attainment_surfaces/mo_nas_201_cifar100_attainment_surfaces_CI__0C.pdf}
       & \includegraphics[trim=0.8cm 0.5cm 0cm 0cm,clip,width=0.315\textwidth]{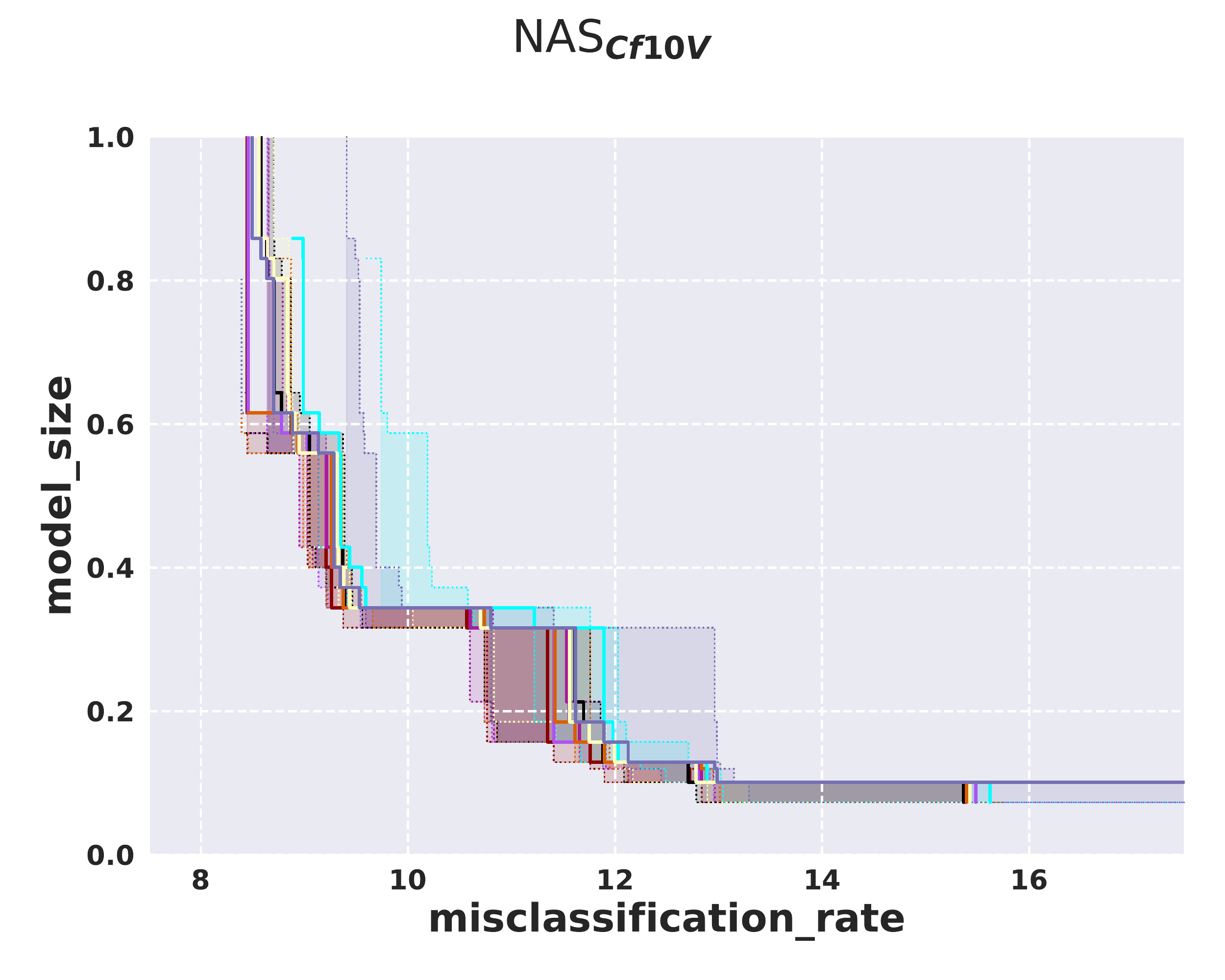}
       & \includegraphics[trim=0.8cm 0.5cm 0cm 0cm,clip,width=0.315\textwidth]{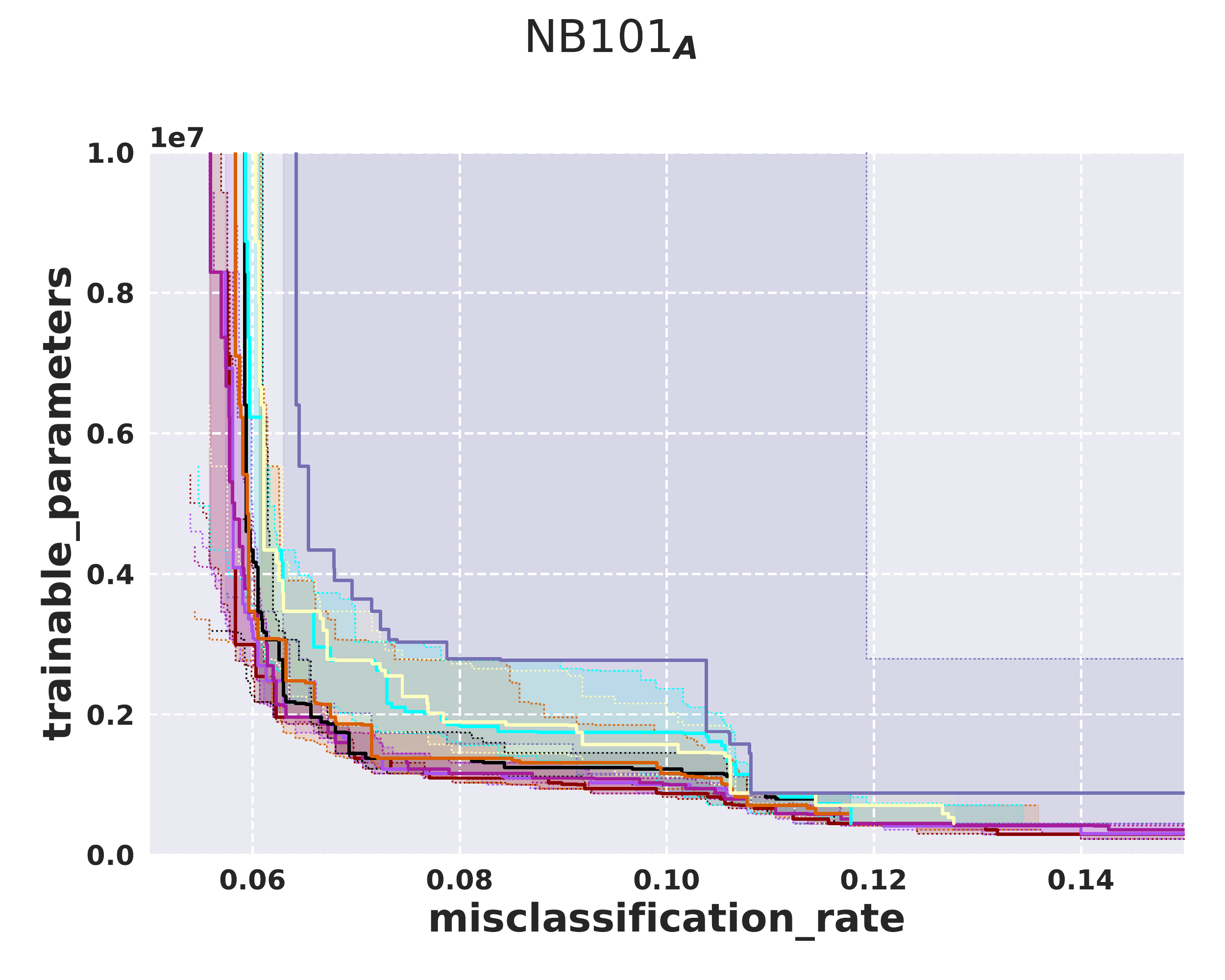}
       & \includegraphics[trim=0.8cm 0.5cm 0cm 0cm,clip,width=0.315\textwidth]{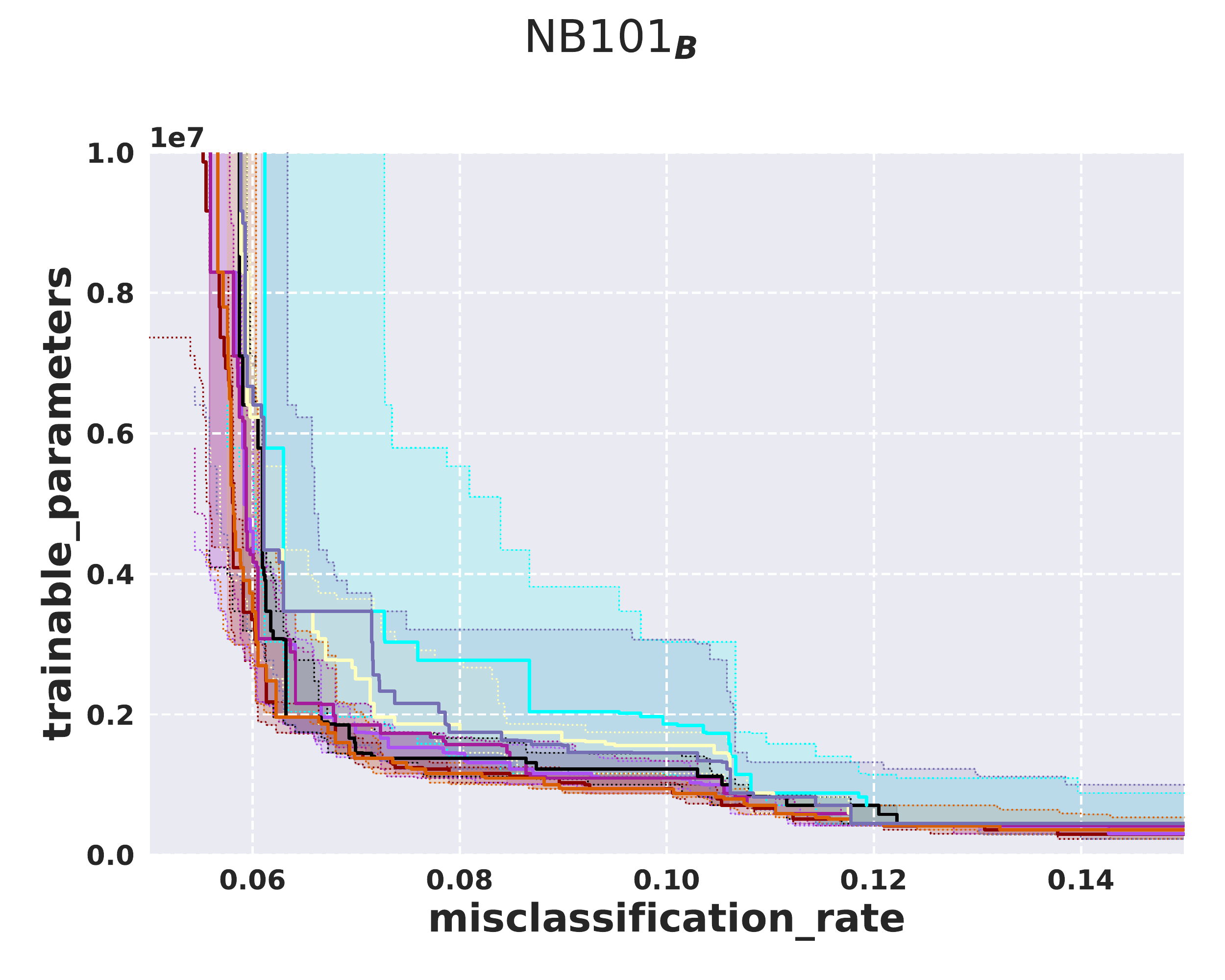}
       & \includegraphics[trim=0.8cm 0.5cm 0cm 0cm,clip,width=0.315\textwidth]{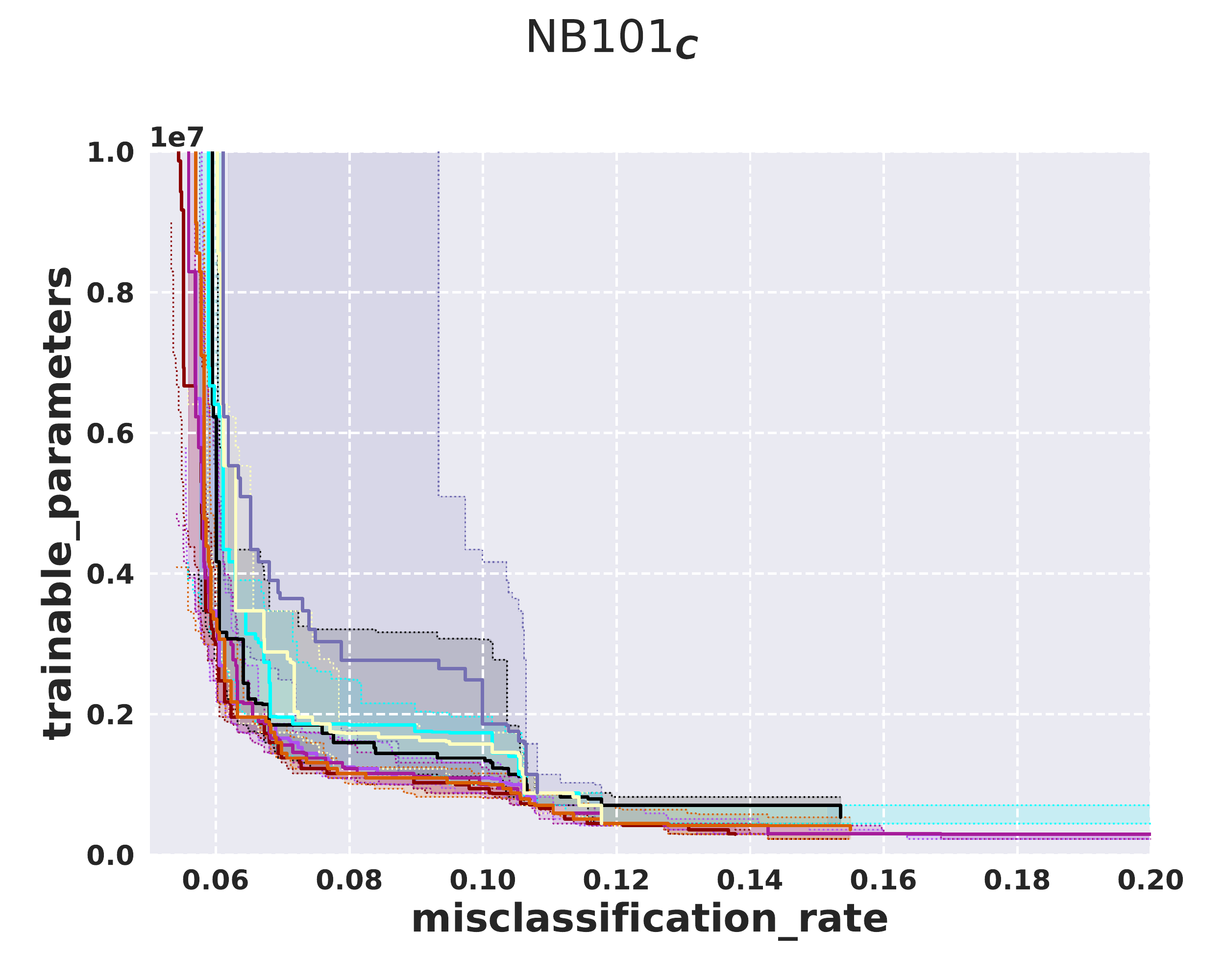}
       & \includegraphics[trim=0.8cm 0.5cm 0cm 0cm,clip,width=0.315\textwidth]{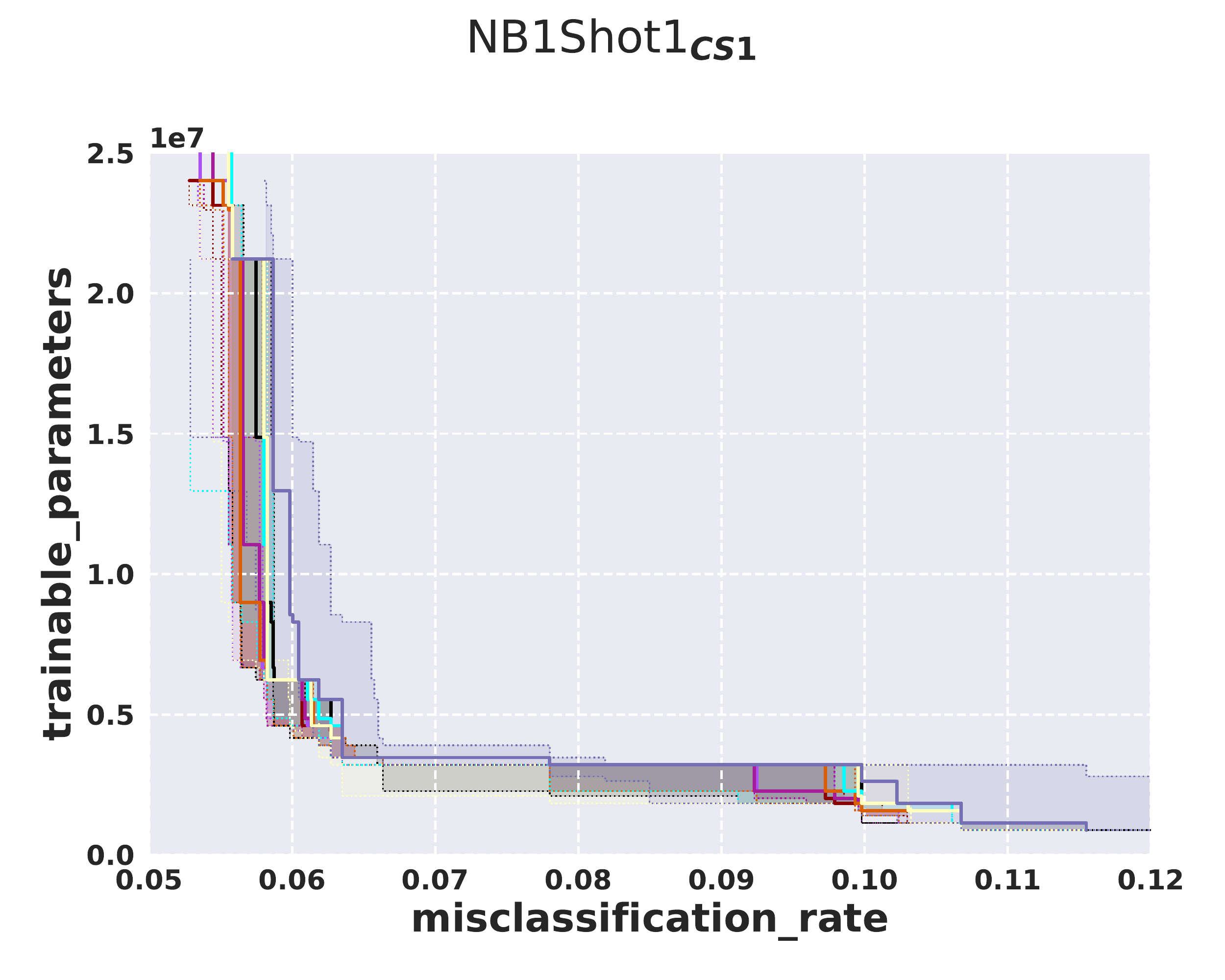}
       & \includegraphics[trim=0.8cm 0.5cm 0cm 0cm,clip,width=0.315\textwidth]{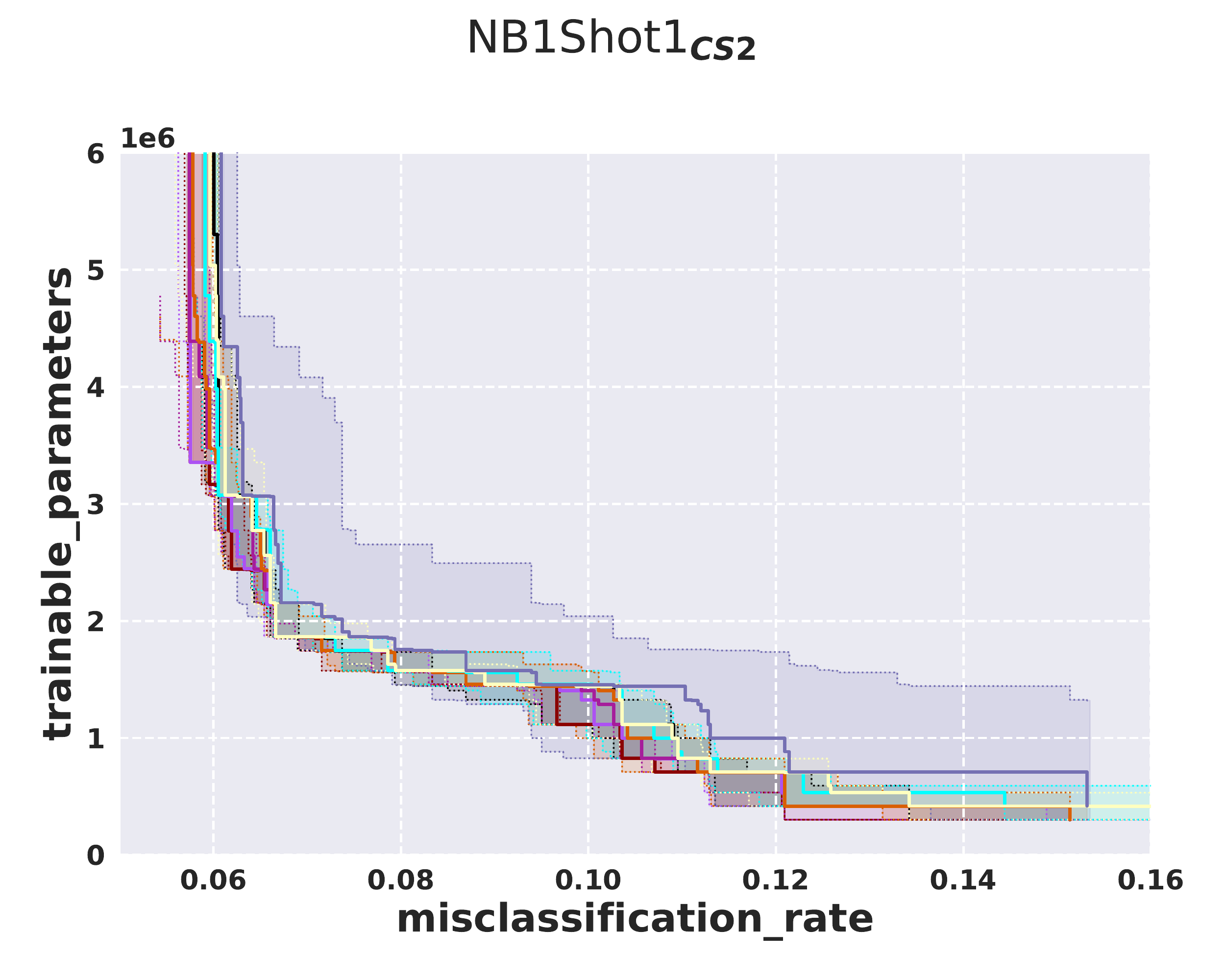}
       & \includegraphics[trim=0.8cm 0.5cm 0cm 0cm,clip,width=0.315\textwidth]{plots/attainment_surfaces/mo_jahs_color_histology_attainment_surfaces_CI__1C.pdf}
       & \includegraphics[trim=0.8cm 1cm 0cm   0cm,clip,width=0.315\textwidth]{plots/attainment_surfaces/mo_jahs_fashion_attainment_surfaces_CI__1C.pdf}
       & \includegraphics[trim=0.8cm 1cm 0cm   0cm,clip,width=0.315\textwidth]{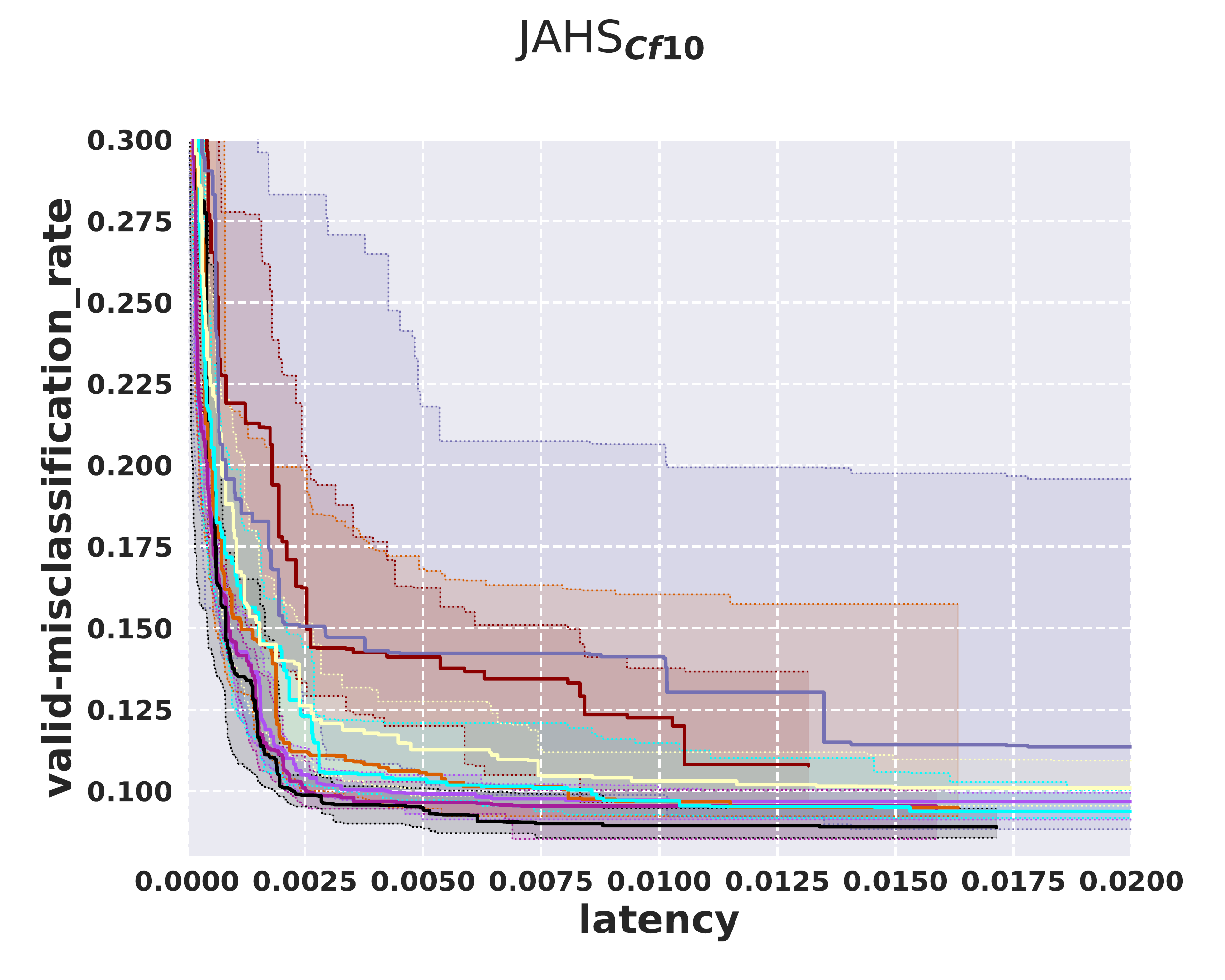}
       & \includegraphics[trim=0     1cm 0cm   0cm,clip,width=0.33\textwidth]{plots/attainment_surfaces/mo_adult_attainment_surfaces_CI__1C.pdf}
       & \includegraphics[trim=0     1cm 0cm   0cm,clip,width=0.33\textwidth]{plots/attainment_surfaces/mo_cnn_flower_attainment_surfaces_CI__0C.pdf}
       & \includegraphics[trim=0     1cm 0cm   0cm,clip,width=0.33\textwidth]{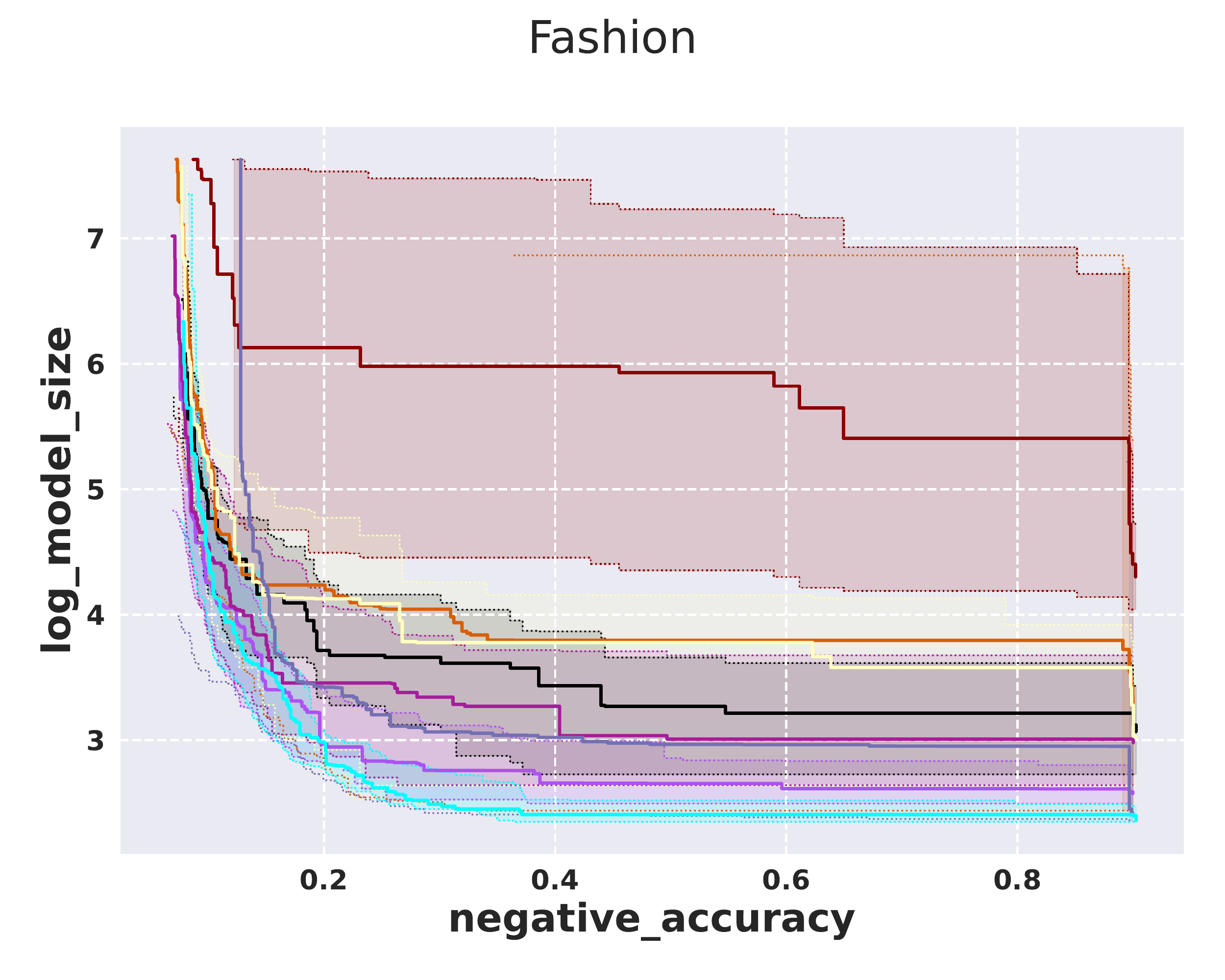}
  \end{tabular}
  \vspace{0cm}
  \includegraphics[trim=0.9cm 0.5cm 0.5cm 0cm,clip, width=0.8\textwidth]{plots/ranking_plots/Ranking_ijcai_all_ijcai_ff_100_legend.pdf}
  \end{center}
  \caption{We report summary-attainment-surfaces for all benchmarks. Upper and lower bound correspond to the first and ninth summary-attainment-surface.}
  \label{fig:attainment_surfaces}
\end{figure*}

In Figure~\ref{fig:attainment_surfaces}, we visualize the summary of attainment surfaces to evaluate the capacity of the baselines methods to approximate the entire Pareto front~\cite{knowles2005summary}. To facilitate the visual inspection of the differences, we show the first, median and ninth attainment surfaces, rather than plotting all 10 attainment surfaces.
In Figure~\ref{fig:attainment_surfaces}, we observe that almost all baselines perform competitively on NAS-Bench-201 and NAS-Bench-1Shot1. For NAS-Bench-101, evolutionary algorithms (EAs) methods (i.e. \agemoead{}, \nsgaiii{} and our \modehb{} variants) perform quite competitively. Moreover, we observe that while SMAC performs better than \modehb{} variants on JAHS-Bench, \modehb{} variants still exhibit consistent and good performance. For fashion dataset, we see \qnparego{} showing better performance than \modehb{} while both variants of \modehb{} shows better performance than other baselines. 
Additionally, we observe that \modehb{} consistently shows superior performance on the Adult dataset. In conclusion, we observe that \modehb{} consistently demonstrates strong performance on all benchmarks.

\bibliographystyle{named}
\bibliography{bib/shortstrings,bib/lib,bib/local,bib/shortproc}

%% file: macros.tex
\usepackage{todonotes}
\presetkeys{todonotes}{inline,backgroundcolor=red!40,caption={}}{}

\usepackage[linesnumbered, ruled]{algorithm2e}
\usepackage{amssymb,dsfont,amsthm,amsmath}

\newcommand{\R}{\mathds{R}}

\newcommand{\mo}{multi-objective}

\newcommand{\hpo}{hyperparameter optimization}
\newcommand{\nas}{Neural Architecture Search}

\newcommand{\hb}{HB}
\newcommand{\sh}{SH}

\newcommand{\tpe}{TPE}

\newcommand{\dehb}{DEHB}
\newcommand{\de}{DE}
\newcommand{\random}{RS}
\newcommand{\modehb}{MO-DEHB}

\newcommand{\modehbepsnet}{MO-DEHB$_{EPSNET}$}
\newcommand{\modehbnsgaii}{MO-DEHB$_{NSGA-II}$}

\newcommand{\qnparego}{QNPAREGO}
\newcommand{\nsgaiii}{NSGA-III}
\newcommand{\moead}{MOEA/D}
\newcommand{\agemoead}{AGE-MOEA}
\newcommand{\RS}{RS}

\newcommand{\hpobench}{HPOBench}